\documentclass{article}
\usepackage{latexsym, amssymb, amsfonts}
\usepackage[dvips]{graphicx}
\usepackage{fancyvrb}

\newcommand*{\implies}{\mathbin{\Rightarrow}}

\title{Probabilistic Reasoning as Information Compression
by Multiple Alignment, Unification and Search: An Introduction and
Overview\footnote{Published in the {\em Journal of Universal Computer Science}
{\bf 5} (7), 418--462, 1999.}}

\author{J Gerard Wolff\\
\small (University of Wales, Bangor, UK\\
gerry@sees.bangor.ac.uk)}

\begin{document}

\maketitle

\begin{abstract}
\sloppy This article introduces the idea that probabilistic reasoning (PR) may
be understood as {\it information compression by multiple alignment,
unification and search} (ICMAUS). In this context, multiple alignment has a
meaning which is similar to but distinct from its meaning in bio-informatics,
while unification means a simple merging of matching patterns, a meaning which
is related to but simpler than the meaning of that term in logic.

A software model, SP61, has been developed for the discovery and formation of
`good' multiple alignments, evaluated in terms of information compression. The
model is described in outline.

Using examples from the SP61 model, this article describes in outline how the
ICMAUS framework can model various kinds of PR including: PR in best-match
pattern recognition and information retrieval; one-step `deductive' and
`abductive' PR; inheritance of attributes in a class hierarchy; chains of
reasoning (probabilistic decision networks and decision trees, and PR with
`rules'); geometric analogy problems; nonmonotonic reasoning and reasoning with
default values; modelling the function of a Bayesian network.
\end{abstract}


{\bf Category:} SD I.2.3.

\section{Introduction}\label{intro}
Quoting Benjamin Franklin (``Nothing is certain but death and taxes''),
Ginsberg \cite[p. 2]{r17} writes that: ``The view that Franklin was expressing
is that virtually every conclusion we draw [in reasoning] is an uncertain
one.'' He goes on to say: ``This sort of reasoning in the face of uncertainty
... has ... proved to be remarkably difficult to formalise.''

This article introduces the idea that {\it probabilistic reasoning} (PR) may be
understood as a process of {\it information compression} (IC) by {\it multiple
alignment} with {\it unification} and {\it search} (ICMAUS). The article is
intended as a summary or overview of research which is described in more
details elsewhere \cite{r38, r39, r40}. In the space available, it is only
possible to present a sketch of the main ideas. Many details are omitted and
there is only brief discussion of assumptions and related issues.

In the ICMAUS framework, {\it multiple alignment} (MA) has a meaning which is
similar to but distinct from its meaning in bio-informatics while {\it
unification} means a simple merging of matching patterns, a meaning which is
related to but simpler than the meaning of that term in logic. The term search
in this context means the systematic exploration of the abstract space of
possible alignments, normally constrained in some way (using heuristic
techniques or otherwise) to achieve useful results in realistic timescales.

In this article, the way in which the IC associated with any alignment may be
calculated is described in outline together with a brief description of SP61, a
software model designed to discover and construct MAs which are `good' in terms
of IC. More detail may be found in \cite{r38, r39, r40}.

With examples from the SP61 model, the main body of the article presents an
overview of how the ICMAUS framework can accommodate a variety of kinds of PR
including:
\begin{itemize}
\item Best-match pattern recognition and information retrieval.

\item Inheritance of attributes in a class hierarchy.

\item One-step `deductive' and `abductive' reasoning.

\item Chains of reasoning:
\begin{itemize}
\item Reasoning with probabilistic decision networks and decision trees.

\item Reasoning with `rules'.
\end{itemize}
\item Reasoning with default values.

\item Nonmonotonic reasoning.

\item Solving geometric analogy problems.

\item ICMAUS as a possible alternative to Bayesian networks.
\end{itemize}

Topics which are discussed in \cite{r39} but are omitted from this article
include: recognition of patterns with internal structure (illustrated with an
example of medical diagnosis); multiple inheritance; and the recognition of
polythetic categories. Topics which are discussed in \cite{r40} but omitted
here include: modelling of `variables' with `values' and `types'; hypothetical
(``what if'') reasoning; indirection in information retrieval; and the
representation of knowledge in the ICMAUS framework.

For the sake of clarity and to save space, the examples presented in Section
\ref{patt-recognition} and the following sections are relatively small.
However, the SP61 model is capable of handling more complicated examples as can
be seen in \cite{r39, r40, r41a}. The scaling properties of the model are good
(see Section \ref{comp-complexity}).

\subsection{Background and context}\label{background}

The proposals in these articles have been developed within a programme of
research developing the `SP' conjecture that:

\begin{quote}
{\it All kinds of computing and formal reasoning may usefully be understood as
information compression by multiple alignment, unification and search,}
\end{quote}

\noindent and developing a `new generation' computing system based on this
thinking.  \footnote{IC may be interpreted as a process of removing unnecessary
(redundant) complexity in information - and thus maximising {\it simplicity} -
whilst preserving as much as possible of its non-redundant descriptive {\it
power}. Hence the name `SP' applied to the central conjecture and other aspects
of this research.}

Background thinking for this research programme is described in \cite{r49} and
\cite{r44}. In addition to PR, the concepts have so far been developed in
relation to the following fields: {\it best-match information retrieval and
pattern recognition} \cite{r47}; {\it parsing of natural language} \cite{r41a};
and {\it automation of software design and the execution of software functions}
\cite{r48}.

Although the ICMAUS framework has not yet been developed for learning, the
entire programme of research is based on earlier research on {\it unsupervised
inductive learning} \cite{r50, r51, r52} which is itself based on principles of
Minimum Length Encoding (MLE\footnote{MLE is used here as an umbrella term for
Minimum Message Length encoding and Minimum Description Length encoding.}, see
\cite{r7, r25, r27, r32, r36, r22}). A preliminary account of the ICMAUS
framework and its range of applications in learning and reasoning was presented
in \cite{r43} at a stage before a working model had been developed or the
concepts had been quantified.

It has been argued \cite{r42} that the ICMAUS framework provides an
interpretation in terms of IC of the Post Canonical System and Universal Turing
Machine models of computing.

\subsubsection{Research on probabilistic reasoning}\label{research-on}

There is now a huge literature on PR and related ideas ranging over `standard'
parametric and non-parametric statistics; {\it ad hoc} uncertainty measures in
early expert systems; Bayesian statistics; Bayesian/belief/causal networks;
Markov networks; Self-Organising Feature Maps; fuzzy set theory and `soft'
computing'; the Dempster-Shaffer theory; abductive reasoning; nonmonotonic
reasoning and reasoning with default values; autoepistemic logic, defeasible
logic, probabilistic, possibilistic and other kinds of logic designed to
accommodate uncertainty; MLE; algorithmic probability and algorithmic
complexity theory; truth maintenance systems; decision analysis; utility
theory; and so on.

A well-known authoritative survey of the field, with an emphasis on Bayesian
networks, is provided by Judea Pearl \cite{r24} although this book is now,
perhaps, in need of some updating. A useful review from the same year is
\cite{r18}.

\sloppy A more recent collection of articles, which together provide a broad
coverage of the subject, appears in \cite{r14}. A relatively short but useful
review of ``uncertainty handling formalisms'' is provided by \cite{r23}.
Regarding the application of different kinds of `logic' to nonmonotonic and
uncertain reasoning, there is a mine of useful information in the articles in
\cite{r14} covering such things as `default logic', `autoepistemic logic',
`circumscription', `defeasible logic', `uncertainty logics' and `possibilistic
logic'. In that volume, the chapter by Ginsberg \cite{r17} provides an
excellent introduction to the problems of nonmonotonic reasoning.

Papers by \cite{r9, r5, r20, r21, r30, r31} are also relevant as are the papers
in \cite{r15}.

\subsubsection{Information compression and probabilistic reasoning}\label{info-compress}

Naturally enough, much of the literature on probabilistic reasoning deals
directly with concepts of probability, especially conditional probability.
Since, however, there is a close connection between probability and compression
(mediated by coding schemes such as the Huffman coding scheme - see, for
example, \cite{r8} - or the Shannon-Fano-Elias coding scheme, {\it ibid}.),
concepts of probability imply corresponding concepts of compression.

That said, a primary emphasis on compression rather than probability provides
an alternative perspective on the subject which may prove useful. Relevant
sources include \cite{r9, r10, r18, r19, r22, r30, r35} and \cite{r37}.

\section{Multiple alignment problems}\label{mult-align-probs}

The term {\it multiple alignment} is normally associated with the computational
analysis of (symbolic representations of) sequences of DNA bases or sequences
of amino acid residues as part of the process of elucidating the structure,
functions or evolution of the corresponding molecules. The aim of the
computation is to find one or more alignments of matching symbols in two or
more sequences which are, in some sense, `good'. Possible meanings for that
term are discussed in Section \ref{eval-align}, below. An example of an
alignment of DNA sequences is shown in Figure \ref{DNA}.

\begin{figure}
\centering
\begin{BVerbatim}
  G G A     G     C A G G G A G G A     T G     G   G G A
  | | |     |     | | | | | | | | |     | |     |   | | |
  G G | G   G C C C A G G G A G G A     | G G C G   G G A
  | | |     | | | | | | | | | | | |     | |     |   | | |
A | G A C T G C C C A G G G | G G | G C T G     G A | G A
  | | |           | | | | | | | | |   |   |     |   | | |
  G G A A         | A G G G A G G A   | A G     G   G G A
  | |   |         | | | | | | | |     |   |     |   | | |
  G G C A         C A G G G A G G     C   G     G   G G A
\end{BVerbatim}
\caption{\small A `good' alignment amongst five DNA sequences (adapted from
Fig. 6 in \protect\cite{r28}, with permission from Oxford University Press).}
\label{DNA} \normalsize
\end{figure}

In this area of research, it is widely recognised that the number of possible
alignments of symbols is normally too large to be searched exhaustively and
that, to achieve a search which has acceptable speed and acceptable scaling
properties, `heuristic' techniques must normally be used. Heuristic techniques
include `hill climbing' (sometimes called `descent'), `beam search', `genetic
algorithms', `simulated annealing', `dynamic programming' and others. With
these techniques, searching is done in stages, with a progressive narrowing of
the search in successive stages using some kind of measure of goodness of
alignments to guide the search. These techniques may be described generically
as `metrics-guided search'.

With these techniques, ideal solutions cannot normally be guaranteed but
acceptably good approximate solutions can normally be found without excessive
computational demands.

There is now a fairly large literature about methods for finding good
alignments amongst two or more sequences of symbols. Some of the existing
methods are reviewed in \cite{r3, r6, r11, r34}.

Because of the way in which the concept of MA has been generalised in this
research (see next), none of the current methods for finding MAs are suitable
for incorporation in the proposed SP system. Hence the development of a new
method, outlined in Section \ref{generalisation}.

\subsection{Generalisation of the concept of multiple alignment}\label{generalisation}

In this research, the concept of MA has been generalised in the following way:

\begin{enumerate}
\item One (or more) of the sequences of symbols to be aligned has a special
status and is designated as `New'. The way in which the concept of `New'
appears to relate to established concepts in computing is shown in Table
\ref{NEW-OLD}.

\item All other sequences are designated as `Old'. The way in which the concept
of `Old' appears to relate to established concepts in computing is also shown
in Table \ref{NEW-OLD}.

\item A `good' alignment is one which, through the unification of symbols in
New with symbols in Old, and through unifications amongst the symbols in Old,
leads to a relatively large amount of compression of New in terms of the
sequences in Old. How this may be done is explained in Section
\ref{eval-align}, below.

\item An implication of this way of framing the alignment problem is that, by
contrast with `multiple alignment' as normally understood in bio-informatics,
any given sequence in Old may appear two or more times in any one alignment and
may therefore be aligned with itself (with the obvious restriction that any one
instance of a symbol may {\bf not} be aligned with itself).\footnote{With the
kind of MA shown in Figure \ref{DNA}, it is obviously possible to include two
or more {\it copies} of a given sequence in any one alignment. To my knowledge,
this is never done in practice because it would simply lead to the trivial
alignment of each symbol in one copy with the corresponding symbol or symbols
in one or more other copies. What is proposed for the ICMAUS framework is
different: any {\it one} pattern may appear two or more times in an alignment.
Each appearance is just that - it is an {\it appearance} of {\it one} pattern,
not a duplicate {\it copy} of a pattern. Since each appearance of a pattern
represents the same pattern, it makes no sense to match a symbol from one
appearance with the corresponding symbol in another appearance because this is
simply matching one instance of the symbol with itself. Any such match is
spurious and must be forbidden.}
\end{enumerate}

\begin{table}
\centering
\begin{tabular}{l l l}
\it Area of & \it New & \it Old \\
\it application \\
\\
Unsupervised &  `Raw' data. &      Grammar or other \\
inductive learning \space \space & &    knowledge structure \\
 & &                    created by learning. \\
\\
Parsing &   The sentence  &  The grammar used for \\
 &    to be parsed. &    parsing. \\
\\
Pattern &   A pattern to be &   The stored knowledge \\
recognition &   recognised &        used to recognise \\
and scene & or scene to be &    one pattern or several \\
analysis &  analysed. &  within a scene. \\
\\
Databases & A `query' in SQL or &   Records stored \\
 &    other query language. \space \space & in the database. \\
\\
Expert &    A `query' in the &  The `rules' or other \\
system &    query language for &    knowledge stored in \\
 &    the expert system. &  the expert system. \\
\\
Computer &  The `data' or &  The computer program \\
program &   `parameters' &    itself. \\
 &    supplied to the \\
 &    program on each run. \\
\end{tabular}
\caption{\small The way in which the concepts of `New' and `Old' in
this research appear to relate to established concepts in computing.}
\label{NEW-OLD} \normalsize
\end{table}

It should be clear that this concept of MA (and the bio-informatics version of
the concept) may be generalised to two-dimensional (or even higher-dimensional)
patterns. There is likely to be a case, at some stage in the SP research
programme, for extending the ideas described in this article into the domain of
two or more dimensions.

\section{The ICMAUS framework}\label{framework}

The main concepts to be presented can probably best be described with reference
to a simple example. Since `parsing' in the sense understood in theoretical
linguistics and natural language processing has come to be a paradigm for the
several concepts to be described, it will provide our first example despite the
fact that, when the input sentence or phrase to be parsed is complete, there is
no significant PR as understood in this article. Sections 3.1 and 3.2 show how
the parsing of a {\it very} simple sentence with a {\it very} simple grammar
may be understood as MA. Much more elaborate examples can be found in
\cite{r41a}.

\subsection{Representing a grammar with patterns of symbols}\label{repr-grammar}

Figure \ref{CF-PSG} shows a simple context-free phrase-structure grammar
(CF-PSG) describing a fragment of the syntax of English. This grammar generates
the four sentences `j o h n r u n s', `j o h n w a l k s', `s u s a n r u n s',
and `s u s a n w a l k s'. Any of these sentences may be parsed in terms of the
grammar, giving a labelled bracketing like this:

\begin{center}
\begin{BVerbatim}
(S(N j o h n)(V r u n s))
\end{BVerbatim}
\end{center}

or an equivalent representation in the form of a tree.

\begin{figure}
\centering
\begin{BVerbatim}
S -> N V
N -> j o h n
N -> s u s a n
V -> w a l k s
V -> r u n s
\end{BVerbatim}
\caption{\small A CF-PSG describing a fragment of English syntax.}
\label{CF-PSG} \normalsize
\end{figure}

Figure \ref{GRAMMAR-PATTERNS} shows the grammar from Figure \ref{CF-PSG}
expressed as a set of {\it strings}, {\it sequences} or {\it
patterns}\footnote{In this programme of research, the term {\it pattern} means
an array of symbols in one {\it or more} dimensions. This includes arrays in
two or more dimensions as well as one-dimensional sequences. Although
one-dimensional sequences will be the main focus of our attention in this
article, the term {\it pattern} will be used as a reminder that the concept of
multiple alignment in this research includes alignments of patterns in two or
more dimensions. Formal definitions of terms like {\it pattern} and {\it
symbol} are provided in Appendix A1 of \cite{r38}.} of symbols. Each pattern in
this `grammar' is like a re-write rule in the CF-PSG notation except that the
rewrite arrow has been removed, some other symbols have been introduced (`0',
`1' and symbols with an initial `\#' character) and there is a number to the
right of each rule.\footnote{For the remainder of this article, quote marks
will be dropped when referring to any grammar like that in Figure
\ref{GRAMMAR-PATTERNS} which is expressed as patterns of symbols. Likewise, the
word `rule' with respect to this kind of grammar will be referred to without
quote marks.},\footnote{This example of a grammar and how it is used in
parsing may give the impression that the ICMAUS framework is merely a trivial
variation of familiar concepts of context-free phrase-structure grammar
(CF-PSG) with their well-known inadequacies for representing and analysing the
`context sensitive' structures found in natural languages. The examples
presented in \cite{r41a} show that the ICMAUS framework is much more `powerful'
than CF-PSGs and can accommodate quite subtle context-sensitive features of
natural language syntax in a simple and elegant manner. Achieving this
expressive power with a relatively simple notation is made possible by the
relatively sophisticated search processes which lie at the heart of the SP
model.},\footnote{For the sake of clarity in exposition and to save space, all
the grammars shown in this article are much simpler than in any practical
system. For similar reasons, all examples of MAs which are presented have been
chosen so that they are small enough to fit on one page without resorting to
font sizes which are too small to read. However, for the reasons given in
Section \ref{comp-complexity}, the model appears to be general and scalable to
realistically large knowledge structures and alignments.}

The number to the right of each rule in Figure \ref{GRAMMAR-PATTERNS} is an
imaginary frequency of occurrence of the rule in a parsing of a notional sample
of the language. These frequencies of occurrence will be discussed later.

The reasons for the symbols which have been added to each rule will become
clear but a few words of explanation are in order here. The symbols `0' and `1'
have been introduced to differentiate the two versions of the `N' patterns and
the two versions of the `V' patterns. They enter into matching and unification
in exactly the same way as other symbols. Although the symbols are the same as
are used in other contexts to represent numbers they do not have the meaning of
numbers in this grammar.

\begin{figure}
\centering
\begin{BVerbatim}
S N #N V #V #S (1000)
N 0 j o h n #N (300)
N 1 s u s a n #N (700)
V 0 w a l k s #V (650)
V 1 r u n s #V (350)
\end{BVerbatim}
\caption{\small A simple grammar written as patterns of symbols. For each rule,
there is a number on the right representing the frequency of occurrence of the
rule in a notional sample of the language.} \label{GRAMMAR-PATTERNS}
\normalsize
\end{figure}

The symbols which begin with `\#' (e.g., `\#S', `\#NP') serve as `termination
markers' for patterns in the grammar. Although their informal description as
`termination markers' suggests that these symbols are meta symbols with special
meaning, they have no hidden meaning and they enter into matching and
unification like every other symbol.

In general, all the symbols which can be seen in Figure \ref{GRAMMAR-PATTERNS}
enter into matching and unification in the same way. Although some of these
symbols can be seen to serve a distinctive role, there is no hidden meaning
attached to any of them; and there is no formal distinction between upper- and
lower-case letters or between digit symbols and alphabetic symbols - and so
on.\footnote{The foregoing assertions are not strictly true of the method of
evaluating alignments which is used in the SP61 model. The principle of ``no
meta symbols'' and thus ``no hidden meanings for symbols'' is an ideal which
this research aims to attain. But, as a temporary solution to the problem of
scoring alignments in the SP61 model, pending something better, a distinction
has been recognised between symbols which begin with `\#' and all other symbols
(details of the scoring method are presented in Section 4 of \cite{r38}).}

\subsection{Parsing as an alignment of a sentence and rules in a grammar}\label{parse-as-align}

Figure \ref{PARSING} shows how a parsing of the sentence `j o h n r u n s' may
be seen as an alignment of patterns which includes the sentence pattern and
other patterns from the grammar shown in Figure \ref{GRAMMAR-PATTERNS}.

The similarity between this alignment and the conventional parsing may be seen
if the symbols in the alignment are `projected' on to a single sequence, thus:

\begin{center}
\begin{BVerbatim}
S N 0 j o h n #N V 1 r u n s #V #S.
\end{BVerbatim}
\end{center}

In this projection, the two instances of `N' in the second column of the
alignment have been merged or `unified', and likewise for the two instances of
`\#N' in the eighth column, and so on wherever there are two or more instances
of a symbol in any column.

\begin{figure}
\centering
\begin{BVerbatim}
0       j o h n        r u n s       0
        | | | |        | | | |
1   N 0 j o h n #N     | | | |       1
    |           |      | | | |
2 S N           #N V   | | | | #V #S 2
                   |   | | | | |
3                  V 1 r u n s #V    3
\end{BVerbatim}
\caption{\small Parsing of the sentence `j o h n r u n s' as an alignment
amongst sequences representing the sentence and relevant rules in the grammar
in Figure \ref{GRAMMAR-PATTERNS}.} \label{PARSING} \normalsize
\end{figure}

This projection is the same as the conventional parsing except that `0' and `1'
symbols are included, right bracket symbols (`)') are replaced by `termination
markers' and each left bracket is replaced by an upper-case symbol which may
also be regarded as a `label' for the structure.

As was noted in Section \ref{generalisation}, the sentence or other sequence of
symbols to be parsed is regarded as New, while the rules in the grammar are
regarded as Old. For the sake of readability and ease of interpretation, New is
normally placed at the top of each alignment with patterns from Old below it.

For the sake of clarity in Figure \ref{PARSING} and other alignments shown
later, each appearance of a pattern in any alignment is given a line to itself.
Apart from the convention that New is always at the top, the order in which
patterns appear (from top to bottom of the alignment) is entirely arbitrary. An
alignment in which the patterns appear in one order is entirely equivalent to
an alignment in which they appear in any other order, provided all other
aspects of the alignment are the same.

\subsection{Evaluation of an alignment in terms of IC}\label{eval-align}

What is the difference between a `good' alignment and a `bad' one? Intuitively,
a good alignment is one which has many {\it hits} (positive matches between
symbols), few {\it gaps} (sequences of one or more symbols which are not part
of any hit) and, where there are gaps, they should be as short as possible.

It is possible to use measures like these directly in computer programs for
finding good MAs and, indeed, they commonly are. However, our confidence in the
validity of measures like these may be increased if they can be placed within a
broader theoretical framework. Concepts of information, IC, and related
concepts of probability provide a suitable framework. Work on the evaluation of
MAs in this tradition includes \cite{r1, r2, r6, r13, r26, r47}.

As was indicated in Section \ref{generalisation}, {\it a good alignment is
defined here as one which provides a basis for an economical coding of New in
terms of the patterns Old}. There is no space here to describe in detail the
method of evaluation which is used in the ICMAUS framework and in SP61. The
outline description here should give readers an intuitive grasp of the method.
A much fuller description may be found in Section 4 of \cite{r38}.

At the most fine-grained level in an alignment like the one shown in Figure
\ref{PARSING}, individual symbols may be encoded using Huffman coding or
Shannon-Fano-Elias (S-F-E) coding (see \cite{r8}) to take advantage of
variations in the frequencies of symbol types.

For individual words in New, patterns like those shown in Figure
\ref{GRAMMAR-PATTERNS} provide suitable codes. Thus, for the example shown in
Figure \ref{PARSING}, `j o h n' may be encoded with the symbols `N 0 \#N' and
`r u n s' may be encoded with `V 1 \#V'. If `j o h n r u n s' were encoded
purely at the level of words, the result would be `N 0 \#N V 1 \#V' which is
not much shorter (in numbers of symbols) than the original sentence. However,
we can take advantage of the fact that within the pattern `N 0 \#N V 1 \#V',
there is a subsequence `N \#N V \#V' and this subsequence is encoded at a
`higher' level by the pattern for a whole sentence, `S N \#N V \#V \#S'.

In order to exploit this higher level encoding, we first add the symbols `S'
and `\#S' (representing the sentence pattern) to the word-level encoding giving
`S N 0 \#N V 1 \#V \#S'. Then we extract the subsequence `N \#N V \#V' (which
is implied by the symbols `S' and `\#S', representing a sentence) so that the
net result is `S 0 1 \#S'. It should be clear that, in the context of the
grammar shown in Figure \ref{GRAMMAR-PATTERNS}, these four symbols
unambiguously encode the original sentence, `j o h n r u n s'. In terms of
numbers of symbols (and in terms of the numbers of bits, calculated by SP61),
this is substantially shorter than the original.

This idea of using `higher level' patterns to encode the codes from `lower
level' patterns may be applied recursively through any number of levels. More
realistic examples may be found in \cite{r41a}.

The method used in SP61 for calculating a measure of the compression associated
with any alignment delivers a value called a {\it compression difference} (CD).
This is the size of New without any compression minus the size of New after it
has been encoded in terms of the alignment.

\subsection{The SP61 model}\label{SP61}

There is no space here to present anything more than a general description of
the SP61 model. A summary of the organisation of the model is presented in
\cite{r38}. SP61 is similar to the SP52 model described quite fully in
\cite{r41a}. The main difference between SP61 and the earlier model is that
SP61 has been generalised to calculate probabilities of inferences. In
addition, a fairly large number of minor refinements have been made.

At the heart of both models is a process for finding alignments between two
patterns which are `good' in terms of IC. An early version of this method for
aligning two patterns is described in \cite{r45} and a more refined version in
\cite{r47}.

The process may be regarded as a form of dynamic programming (see, for example,
\cite{r29}) but it differs from standard methods in three main ways:

\begin{itemize}
\item It can find alternative alignments between two patterns, graded in terms
of compression. \item It exploits list processing techniques so that
arbitrarily long patterns may be compared (within the limits of the host
machine). \item The thoroughness of searching, and thus the computational
resources which are required (computing time or memory size or both), may be
controlled with parameters.
\end{itemize}

For each pattern in New (and there is normally only one), this process is
applied initially to compare the pattern in New with all the patterns in Old.
From amongst the alignments which are found, the best ones are `unified' to
convert the alignment into a single sequence or pattern of symbols. This
pattern is stored together with the alignment from which it was derived. Any
alignments which cannot be unified in this way are discarded.

For each pattern amongst the best of the unified alignments just formed, the
process is repeated, the best alignments are unified to form simple patterns
and these patterns (with their alignments) are stored as before. This cycle is
repeated in the same way until no more alignments can be found.

\subsubsection{Computational complexity}\label{comp-complexity}

Given that all the example grammars in this article are smaller than would be
required in any realistic system, and given the well-known computational
demands of multiple alignment problems, readers may reasonably ask whether the
proposed framework for parsing would scale up to handle realistically large
knowledge structures and patterns in New.

The time complexity of the SP52 model in a serial processing environment has
been estimated to be approximately O$(log_2n \cdot nm)$, where $n$ is the
length of the sentence and $m$ is the sum of the lengths of the patterns in Old
\cite{r41a}. The same estimate is valid for the SP61 model. In a parallel
processing environment, the time complexity of the models may approach
O$(log_2n \cdot n)$, depending on how the parallel processing is applied. In
serial or parallel environments, the space complexity should be O$(m)$.

In summary, there is reason to think that the method of forming alignments
which is embodied in SP61 will not lead to running times or demands for storage
which are hopelessly impractical when the this approach is applied to
realistically large examples.

\subsubsection{Alignments with one-dimensional patterns}\label{one-dimension}

As was indicated in Section \ref{generalisation}, it is envisaged that the
ICMAUS concepts will, at some stage, be generalised to accommodate patterns in
two or more dimensions. However, at present, the SP52 and SP61 models are
restricted to one-dimensional patterns.

One consequence of this restriction is that it is necessary, with any given
alignment, to be able to `project' it into a single sequence as was shown in
Section \ref{parse-as-align}. This can only be done if the left-right position
of every symbol in the alignment is unambiguous. With alignments like these:

\begin{center}
\begin{BVerbatim}
a   b       a b 
|   |       | | 
a x b  and  a b x 
|   |       | | 
a y b       a b y
\end{BVerbatim}
\end{center}

\noindent the relative left-right positions of `x' and `y' are not defined
which means that the alignments cannot be projected into a single sequence. In
the SP52 and SP61 models, all such alignments are `illegal' and are discarded.

When the models are generalised to handle patterns in two (or more) dimensions,
there should be some relaxation in the restriction just described. For example,
if the left-right position of `x' and `y' is undefined in a time dimension, it
should still be possible to accept the alignment provided that `x' and `y' were
at two different positions in a spatial dimension.

Another possible way to avoid this restriction might be to generalise the
models so that, when appropriate, the symbols in each pattern may be treated as
an unordered collection or `bag'. Since order is no longer significant,
alignments like the ones shown above should be legal. No attempt has yet been
made to generalise the models in this way.

\section{Probabilistic reasoning, multiple alignment and information compression}\label{prob-reason}

What connection is there between the formation of an alignment, as described in
Section \ref{parse-as-align}, and PR? This section describes the connection and
describes in outline how the probabilities of inferences may be calculated. A
fuller presentation, with more discussion of the method and the suppositions on
which it is based, may be found in \cite{r39}.

In the simplest terms, PR arises from partial pattern recognition: if a pattern
is recognised from a subset of its parts (something that humans and animals are
very good at doing), then, in effect, an inference is made that the unseen part
or parts are `really' there. We might, for example, recognise a car from seeing
only the front half because the rear half is hidden behind another car or a
building. The inference that the rear half is present is probabilistic because
there is always a possibility that the rear half is absent or, in some surreal
world, replaced by the front half of a horse, and so on.

In terms of multiple alignment, PR may be understood as the formation of an
alignment in which some part or parts of the patterns from Old which appear in
the alignment (single symbols, substrings or subsequences within patterns from
Old) are not aligned with any matching symbol or symbols from New. As a working
hypothesis, all kinds of PR may be understood in these terms.

What connection is there between the probability of any inferences which may
appear in an alignment and measures of compression for that alignment? How can
the probability of MA inferences be calculated? Answers to these two questions
are presented next. Although the example to be shown is simple, the method to
be outlined is general and may be applied to alignments of any complexity.

\subsection{Absolute probabilities of alignments and inferences}\label{absolute-probs}

Any sequence of $L$ symbols, drawn from an alphabet of $|A|$ symbol types,
represents one point in a set of $N$ points where $N$ is calculated as:

    \[N = |A|^L.\]

\noindent {\it If we assume that the sequence is random or nearly so}, which
means that the N points are equi-probable or nearly so, the probability of any
one point (which represents a sequence of length L) is reasonably close to:

    \[p_{ABS} = |A|^{-L}.\]

\noindent This formula can be used to calculate the probability of a New
sequence in an alignment after it has been compressed in terms of the Old
patterns in the alignment: the formula is applied with the value of $L$ being
the length of the New after encoding. In SP61, the value of $|A|$ is $2$.

If New is not completely matched by symbols in Old, we can use the formula to
calculate the probability of that substring or subsequence within New which is
matched to symbols in Old. This probability may also be regarded as a
probability both of the alignment and of any inferences (symbols from Old which
are not matched to New) within the alignment.

\subsubsection{Is it reasonable to assume that New in encoded form is random or nearly so?}\label{new-random}

Why should we assume that the code for an alignment is a random sequence or
nearly so? In accordance with Algorithmic Information Theory (see, for example,
\cite{r22}), a sequence is random if it is incompressible. If we have reason to
believe that a sequence is incompressible or nearly so, then we may regard it
as random or nearly so.

Generally, we cannot prove for any given body of data that no further
compression is possible. But we may say that, with the methods we are currently
using, and the resources we have applied, no further compression can be
achieved. In short, the assumption that the code for an alignment is random or
nearly so only applies to the best encodings found for a given body of
information in New and must be qualified by the quality and thoroughness of the
search methods which have been used to create the code.

\subsection{Relative probabilities of alignments}\label{rel-probs}

The absolute probabilities of alignments, calculated as described in the last
subsection, are normally very small and not very interesting in themselves.
From the standpoint of practical applications, we are normally interested in
the {\it relative} values of probabilities, not their {\it absolute} values.

A point we may note in passing is that the calculation of relative
probabilities from $p_{ABS}$ will tend to cancel out any general tendency for
values of  $p_{ABS}$ to be too high or too low. Any systematic bias in values
of $p_{ABS}$ should not have much effect on the values which are of most
interest to us.

If we are to compare one alignment and its probability to another alignment and
its probability, \textbf{\it we need to compare like with like}. An alignment
can have a high value for $p_{ABS}$ because it encodes only one or two symbols
from New. It is not reasonable to compare an alignment like that to another
alignment which has a lower value for $p_{ABS}$ but which encodes more symbols
from New. Consequently, the procedure for calculating relative values for
probabilities ($p_{REL}$) is as follows:

\begin{enumerate}
\item For the alignment which has the highest CD (which we shall call the
\textbf{\it reference alignment}), identify the symbols from New which are
encoded by the alignment. We will call these symbols the \textbf{\it reference
set} of symbols in New. \item Compile a {\it reference set} of alignments {\it
which includes the alignment with the highest CD and all other alignments (if
any) which encode exactly the reference set of symbols from New, neither more
nor less}.\footnote{There may be a case for defining the reference set of
alignments as those alignments which encode the reference set of symbols {\it
or any super-set of that set}. It is not clear at present which of those two
definitions is to be preferred.} \item The alignments in the reference set are
examined to find and remove any rows which are redundant in the sense that all
the symbols appearing in a given row also appear in another row in the same
order.\footnote{If Old is well compressed, this kind of redundancy amongst the
rows of an alignment should not appear very often.} Any alignment which, after
editing, matches another alignment in the set is removed from the set. \item
Calculate the sum of the values for $p_{ABS}$ in the reference set of
alignments:

\[p_{A\_SUM} = \sum_{i = 1}^{i = R} p_{ABS_i}\]

where $R$ is the size of the reference set of alignments and $p_{ABS_i}$ is the
value of $p_{ABS}$ for the $i$th alignment in the reference set. \item For each
alignment in the reference set, calculate its relative probability as:
    \[p_{REL_i} = p_{ABS_i} / p_{A\_SUM}.\]
\end{enumerate}

The values of $p_{REL}$ calculated as just described seem to provide an
effective means of comparing the alignments which encode the same set of
symbols from New as the alignment which has the best overall CD.

It is not necessary always to use the alignment with the best CD as the basis
of the reference set of symbols. It may happen that some other set of symbols
from New is the focus of interest. In this case a different reference set of
alignments may be constructed and relative values for those alignments may be
calculated as described above.

\subsection{Relative probabilities of patterns and symbols}\label{rel-probs-patts}

It often happens that a given pattern from Old or a given symbol type within
patterns from Old appears in more than one of the alignments in the reference
set. In cases like these, one would expect the relative probability of the
pattern or symbol type to be higher than if it appeared in only one alignment.
To take account of this kind of situation, SP61 calculates relative
probabilities for individual patterns and symbol types in the following way:

\begin{enumerate}
\item Compile a set of patterns from Old, each of which appears at least once
in the reference set of alignments. \item For each pattern, calculate a value
for its relative probability as the sum of the $p_{REL}$ values for the
alignments in which it appears. If a pattern appears more than once in an
alignment, it is only counted once for that alignment. \item Compile a set of
symbol types which appear anywhere in the patterns identified in 2. \item For
each symbol type identified in 3, calculate its relative probability as the sum
of the relative probabilities of the patterns in which it appears. If it
appears more than once in a given pattern, it is only counted once. With regard
to symbol types, the foregoing applies only to symbol types which do not appear
in New. Any symbol type which appears in New necessarily has a probability of
$1.0$ - because it has been observed, not inferred.
\end{enumerate}

\subsection{A simple example}\label{simple-example}

In order to illustrate the kinds of values which may be calculated for absolute
and relative probabilities, this subsection presents a very simple example: the
inference of `fire' from `smoke'. Here, we shall extend the concept of `smoke'
to include anything, like mist or fog, which looks like smoke. Also, `fire' has
been divided into three categories: the kind of fire used to heat a house or
other building, dangerous fires that need a fire extinguisher or more, and the
kind of fire inside a burning cigarette or pipe. The alignment we are
considering looks like this:

\large
\begin{center}
\begin{BVerbatim}
     smoke
       |
fire smoke
\end{BVerbatim}
\end{center}
\normalsize

Given the small knowledge base shown in Table \ref{ASSOCIATIONS} as Old, and a
pattern containing the single symbol `smoke' as New, the frequency of
occurrence and the `minimum' cost of each symbol type (calculated by SP61 using
the method described in Section 4 of \cite{r41a}) are shown in Table
\ref{SYMBOL-TYPES}.

\begin{table}
\centering
\begin{tabular}{lrr}
\it Patterns & \it Frequency & \it Encoding \\
 & & \space \space \it cost (bits) \\
\\
clouds black rain & 15,000 & 2.93 \\
dangerous fire smoke \space \space & 500 & 7.84 \\
heating fire smoke & 7,000 & 4.03 \\
tobacco fire smoke & 10,000 & 3.52 \\
fog smoke & 2,000 & 5.84 \\
stage smoke & 100 & 10.16 \\
thunder lightning & 5,000 & 4.52 \\
strawberries cream & 1,500 & 6.26 \\
\end{tabular}
\caption{\small A small knowledge base of associations. The middle column shows
an imaginary frequency of occurrence of each pattern in some notional reference
environment. The right-hand column shows the encoding cost of each pattern,
calculated as described in Section 4 of \protect\cite{r41a}.}
\label{ASSOCIATIONS} \normalsize
\end{table}

The encoding cost of each pattern is simply the sum of the minimum costs of
each of the minimum number of symbols from the pattern which are needed to
discriminate the pattern uniquely within Old (as described in Section 4 of
\cite{r41a}). In this example, every pattern can be identified uniquely by its
first symbol. Thus, in all cases, the encoding cost of each pattern (shown to
the right of Table \ref{ASSOCIATIONS}) is the minimum cost of its first symbol.

\begin{table}
\centering
\begin{tabular}{lrr}
 & \it Frequency & \it Minimum \\
 & & \it cost (bits) \\
\\
smoke & 19,601 & 2.55 \\
black & 15,000 & 2.93 \\
clouds & 15,000 & 2.93 \\
cream & 1,500 & 6.26 \\
dangerous & 500 & 7.84 \\
fire & 17,500 & 2.71 \\
fog & 2,000 & 5.84 \\
heating & 7,000 & 4.03 \\
lightning & 5,000 & 4.52 \\
rain & 15,000 & 2.93 \\
stage & 100 & 10.16 \\
strawberries \space \space & 1,500 & 6.26 \\
thunder & 5,000 & 4.52 \\
tobacco & 10,000 & 3.52 \\
\end{tabular}
\caption{\small The frequency of occurrence of each symbol type appearing in
Table \ref{ASSOCIATIONS} (together with New which is a pattern containing the
single symbol `smoke') and the minimum cost of each symbol type calculated by
the method described in Section 4 of \protect\cite{r41a}.} \label{SYMBOL-TYPES}
\normalsize
\end{table}

Given that New is a pattern containing the single symbol `smoke', SP61 forms
the five obvious alignments of New with each of the patterns in Table
\ref{ASSOCIATIONS} which contain the symbol `smoke'. The absolute and relative
probabilities of the five alignments, calculated as described above, are shown
in Table \ref{REL-PROB-ALIGNMENTS}.

\begin{table}
\centering
\begin{tabular}{lll}
& \it Absolute & \it Relative \\
& \it probability & \it probability \\
\\
smoke/tobacco fire smoke & 0.08718 & 0.51020 \\
smoke/heating fire smoke & 0.06103 & 0.35714 \\
smoke/fog smoke & 0.01744 & 0.10204 \\
smoke/dangerous fire smoke & 0.00436 & 0.02551 \\
smoke/stage smoke & 0.00009 & 0.00510 \\
\end{tabular}
\caption{\small Absolute and relative probabilities of each of the five
reference alignments formed between `smoke' in New and, in Old, the patterns
shown in Table \ref{ASSOCIATIONS}. In this example, the relative probability of
each pattern from Old is the same as the alignment in which it appears.}
\label{REL-PROB-ALIGNMENTS} \normalsize
\end{table}

In this very simple example, the relative probability of each pattern from Old
is the same as for the alignment in which it appears. However, the same cannot
be said of individual symbol types. The relative probabilities of the symbol
types that appear in any of the five reference alignments (shown in Table
\ref{REL-PROB-ALIGNMENTS}) are shown in Table 5.

\begin{table}
\centering
\begin{tabular}{ll}
 & \it Relative \\
 & \it probability \\
\\
smoke & 1.00000 \\
fire & 0.89286 \\
tobacco & 0.51020 \\
heating & 0.35714 \\
fog & 0.10204 \\
dangerous & 0.02551 \\
stage & 0.00510 \\
\end{tabular}
\caption{\small The relative probabilities of the symbol types from Old that
appear in any of the reference set of alignments shown in Table
\ref{REL-PROB-ALIGNMENTS}.} \label{REL-PROB-SYMBOL-TYPES} \normalsize
\end{table}

The main points to notice about the relative probabilities shown in Table 5
are:

\begin{itemize}
\item The relative probability of `smoke' is 1.0. This is because it is a
`fact' which appears in New: hence there is no uncertainty attaching to it.
\item Of the other symbol types from Old, the one with the highest probability
relative to the other symbols is `fire', and this relative probability is
higher than the relative probability of any of the patterns from Old (Table
\ref{REL-PROB-ALIGNMENTS}). This is because `fire' appears in three of the
reference alignments.
\end{itemize}

In this example, we have ignored all the subtle cues that people would use in
practice to infer the origin of smoke: the smell, colour and volume of smoke,
associated noises, behaviour of other people, and so on. Allowing for this, and
allowing for the probable inaccuracy of the frequency values which have been
used, the relative probabilities of alignments, patterns and symbols seem to
reflect the subjective probability which we might assign to the five
alternative sources of smoke-like matter in everyday situations.

\section{Best-match pattern recognition and information retrieval}\label{patt-recognition}

As was noted in Section \ref{prob-reason}, recognition of objects or patterns
can entail PR when a pattern is recognised from a subset of its parts. The same
is true of best-match information retrieval. In both cases, there can be errors
of omission, addition and substitution.

As an example of best-match pattern recognition (which may also be construed as
best-match information retrieval), consider the thoroughly mis-spelled `word',
`c m p u x t a r' and how it may be matched against stored patterns. Figure
\ref{SPELLING} shows the four alignments formed by SP61 which have the highest
CDs when this pattern was supplied as New and, in Old, a small dictionary of 45
words prepared by selecting, in a more or less haphazard manner, one, two or
three words from each of the alphabetic sections of an ordinary English
dictionary. Each word was given a notional frequency of occurrence.

\begin{figure}
\centering
\begin{BVerbatim}
0 c   m p u x t a r 0
  |   | | |   |   |
1 c o m p u   t e r 1

(a)

0 c     m p u x t a r 0
  |     |   |   |   |
1 c o m m   u   t e r 1

(b)

0 c   m p u x t a r 0
  |   |   |   |   |
1 c o m m u   t e r 1

(c)

0 c m p u x t a r           0
      |       | |
1     p       a r a f f i n 1

(d)
\end{BVerbatim}
\caption{\small The best four alignments formed by SP61 between `c m p u x t a
r' and words in a small dictionary of 45 words. CD values for the four
alignments are, from the top down, 41.39, 29.14, 28.44 and 13.76.}
\label{SPELLING} \normalsize
\end{figure}

The first three alignments are, by far, the best matches for the given pattern.
They have CDs (from the top alignment downwards) of 41.39, 29.14 and 28.44. The
next best alignment is the fourth one shown in Figure \ref{SPELLING}, which has
a CD of only 13.76. Notice that each of the first three alignments contain
discrepancies between `c m p u x t a r' and the word to which it has been
aligned which represent all three of the possible kinds of discrepancy:
omission, addition and substitution of symbols.

Regarding probabilities, the absolute probability of the best alignment is
calculated by SP61 as 0.00652. Since there is no other alignment which contains
all and only the symbols `c', `m', `p', `u', `t' and `r' from New, the
reference list of alignments containing all and only the same symbols from New
has only one entry, and so the relative probability of the best alignment is
1.0.

In general, these results accord with our intuition that the `correct' form of
`c m p u x t a r' is `c o m p u t e r' and our intuition that `c o m m u t e r'
is a very close alternative.

\section{Inheritance of attributes in a class hierarchy}\label{inheritance}

In describing objects, patterns or other entities it is often useful to assign
them to classes which may themselves be classified recursively through any
number of levels. This, of course, is the basis of object-oriented software and
object-oriented databases. The value of a classification system is that it
saves repetition of features which are the same in two or more entities. Many
animals, for example, have a backbone. If these animals are grouped into a
class `vertebrates' then the feature `backbone' need only be recorded once in
the description of the class and it can then be {\it inherited} by members of
the class without the need to record it in each of the several descriptions of
individual animals. This device is a mechanism for information compression.

Figure \ref{TAXONOMY} shows a set of patterns representing, in highly
simplified form, part of a class hierarchy for vertebrate animals. As in Figure
\ref{GRAMMAR-PATTERNS}, each pattern is followed by a number in brackets which
represents a notional frequency of occurrence of the pattern in some domain.

The first pattern represents the class vertebrates and is mainly a framework
for lower-level classes. It provides empty slots for `name', `head', `legs',
`blood' and others, while the slot for `body' shows that each vertebrate has a
backbone.

\begin{figure}
\centering
\begin{BVerbatim}
vertebrate description name #name head #head
     body backbone #body
     legs #legs blood #blood
     covering #covering food #food
     other_features #other_features
     #description #vertebrate (104700)
vertebrate reptile description blood cold #blood
     covering scaly #covering
     #description #reptile #vertebrate (35000)
vertebrate bird description head beak #head
     blood warm #blood
     covering feathers #covering
     #description #bird #vertebrate (40000)
vertebrate mammal description
     blood warm #blood covering fur #covering
     #description #mammal #vertebrate (29700)
mammal marsupial description
     other_features born_small#other_features
     #description #marsupial #mammal (700)
mammal cetacean description
     other_features
          fish_shape live_in_water
     #other_features
     #description #cetacean #mammal (15000)
mammal carnivore description food flesh_eating #food
     #description #carnivore #mammal (1400)
carnivore cat description
     other_features
          retractile_claws purrs
     #other_features
     #description #cat #carnivore (600)
carnivore dog description
     other_features barks #other_features
     #description #dog #carnivore (800)
\end{BVerbatim}
\caption{\small A set of patterns representing part of a class hierarchy of
vertebrate animals, much simplified. The number in brackets at the end of each
pattern is a notional frequency of occurrence of that pattern.}
\label{TAXONOMY} \normalsize
\end{figure}

Below the pattern for vertebrates - lower on the page and lower in terms of the
class hierarchy - are patterns for reptiles, birds and mammals. Each one is
identified as a vertebrate and each one shows distinctive features for its
class: reptiles are cold-blooded, birds have feathers, and so on. In a similar
way, three subclasses of the class mammal are specified and, within one of
those subclasses (the class `carnivore'), there is a pattern for each of the
classes `dog' and `cat'. In each case, the pattern for a given class is linked
to its `parent' class by inclusion in the pattern of the name of the parent
class.

Figure \ref{RECOGNISE-CAT} shows the best alignment found by SP61 with the
pattern `description name Tibs \#name purrs \#description' in New and, in Old,
the patterns shown in Figure \ref{TAXONOMY}. Each of the symbols from Figure
\ref{TAXONOMY} has been abbreviated in Figure \ref{RECOGNISE-CAT} so that the
alignment does not become too long to be shown easily on a page.

This alignment shows rather clearly how the ICMAUS framework allows inferences
to be drawn. With the patterns shown in Figure \ref{TAXONOMY}, the symbols in
New (particularly the symbol `purrs') imply that the animal is a cat (which
means that it has retractile claws), that it is a carnivore (which means that
it is flesh-eating), that it is a mammal (which means that it is warm blooded
and covered in fur), and that it is a vertebrate and therefore has a backbone.

\begin{figure}
\centering
\begin{BVerbatim}
0             dn ne Ts #ne                                     0
              |  |     |
1 ve ml       dn |     |                           bld wm #bld 1
  |  |        |  |     |                            |      |
2 ve |        dn ne    #ne hd #hd by be #by ls #ls bld    #bld 2
     |        |
3    ml ce    dn                                               3
        |     |
4       ce ct dn                                               4

0                           ps     #dn             0
                            |       |
1 cg fr #cg                 |      #dn     #ml #ve 1
  |      |                  |       |       |   |
2 cg    #cg fd    #fd os    |  #os #dn      |  #ve 2
            |     |   |     |   |   |       |
3           fd fg #fd |     |   |  #dn #ce #ml     3
                      |     |   |   |   |
4                     os rs ps #os #dn #ct #ce     4
\end{BVerbatim}
\caption{\small The best alignment found by SP61 with `description name Tibs
\#name purrs \#description' in New and the patterns shown in Figure
\ref{TAXONOMY} in Old. Names of symbols (shown in Figure \ref{TAXONOMY}) have
been shortened here to reduce the length of the alignment. {\it Key}: be =
backbone, bld = blood, by = body, ce = carnivore, cg = covering, ct = cat, dn =
description, fd = food, fg = flesh\_eating, fr = fur, hd = head, ls = legs, ml
= mammal, ne = name, os = other\_features, ps = purrs, ts = Tibs, ve =
vertebrate.}
\label{RECOGNISE-CAT}
\normalsize
\end{figure}

In this case, there is only one alignment that can match the same symbols from
New. In terms of those symbols, the relative probability of the alignment - and
thus the inferences which can be drawn from the alignment - is 1.0.

No attempt has been made here to show how the ICMAUS framework might
accommodate systems of classes with multiple-inheritance (cross
classification). Readers may like to think how this might be done.

The way in which `polythetic' classes \footnote{A polythetic class is one in
which there need not be any one attribute which is shared by all members of the
class.} may be represented and used in the ICMAUS scheme is discussed in
\cite{r39}.

\subsection{Inferences with less distinctive features}\label{less-distinct-features}

What happens if the symbols in New are not so distinctive for one of the
lowest-level classes? SP61 has been run with the same patterns in Old but with
the pattern `description name Tibs \#name flesh\_eating \#description' in New. In
this case, the program forms three alignments which match with all the symbols
in New (except `Tibs'). The best one in terms of compression is like the
alignment in Figure \ref{RECOGNISE-CAT} except that it does not include a
pattern for the class `cat'. In effect, it identifies the unknown animal as a
carnivore without specifying what kind of carnivore. The other two alignments
are the same except that, in addition, one contains the pattern for `dog' and
the other contains the pattern for `cat'.

The SP61 program calculates relative probabilities for the three alignments as
0.915, 0.062 and 0.023, respectively. From these values, the program calculates
probabilities for individual patterns and symbols in the alignments.

Since the pattern for `carnivore' appears in all three alignments, its
probability is calculated as the sum of the three relative probabilities -
which is 1.0. Likewise for the patterns for `mammal' and `vertebrate'. However,
the pattern for `dog' appears in only one of the three alignments so the
probability is the same as the relative probability for that alignment: 0.062.
In a similar way, the probability for the pattern for `cat' is 0.023. Similar
calculations are made for symbols in the alignments.

The probability value of 1.0 for `carnivore', `mammal' and `vertebrate'
reflects our intuition that, in terms of the patterns in Old, the unknown
animal is certainly a carnivore and also a mammal and a vertebrate. But the
relative probabilities for `dog' and `cat' seem rather low.

A possible refinement of the method of calculating probabilities might be for
the system to recognise that the first alignment is contained within each of
the second and third alignments so that a relative probability may be
calculated for each of the second and third alignments, in each case excluding
the probability of the first alignment. If the calculations are done in this
way, the relative probability of `dog' would be 0.729 and the relative
probability of `cat' would be 0.271. These values accord much better with our
intuitions.

\section{One-step `deductive' reasoning}\label{one-step-reasoning}

Consider a `standard' example of {\it modus ponens} syllogistic reasoning:

\begin{enumerate}
\item $\forall x$: bird($x$) $\implies$ canfly($x$). \item bird(Tweety). \item
$\therefore$ canfly(Tweety).
\end{enumerate}

\noindent which, in English, may be interpreted as:

\begin{enumerate}
\item If something is a bird then that something can fly. \item Tweety is a
bird. \item Therefore, Tweety can fly.
\end{enumerate}

In strict logic, a `material implication' like $(p \implies q)$ (``If something
is a bird then that something can fly'') is equivalent to $(\neg q \implies
\neg p)$ (``If something cannot fly then it is not a bird'') and also to $(\neg
p \implies q)$ (``Either something is not a bird or it can fly'').

However, there is a more relaxed, `everyday' kind of `deduction' which, in
terms of our example, may be expressed as: ``If something is a bird then, {\it
probably}, it can fly. Tweety is a bird. Therefore, {\it probably}, Tweety can
fly.''

This kind of probabilistic `deduction' differs from strict material implication
because it does not have the same equivalencies as the strict form. If our
focus of interest is in describing and reasoning about the real world (rather
than exploring the properties of abstract systems of symbols), the
probabilistic kind of `deduction' seems to be more appropriate. With regard to
birds, we know that there are flightless birds, and for most other examples of
a similar kind, an ``all or nothing'' logical description would not be an
accurate reflection of the facts.

With a pattern of symbols, we may record the fact that birds can fly and, in a
very natural way, we may record all the other attributes of a bird in the same
pattern. The pattern may look something like this:

\begin{center}
\begin{BVerbatim}
bird name #name canfly wings feathers beak
     crop lays_eggs ... #bird
\end{BVerbatim}
\end{center}

\noindent or the attributes of a bird may be described in the more elaborate
way described in Section \ref{inheritance}.

This pattern and others of a similar kind may be stored in `Old', together with
patterns like `name Tweety \#name', `name Tweety \#name', `name Susan \#name' and
so on which define the range of possible names. Also, the pattern, `bird
Tweety', corresponding to the proposition ``Tweety is a bird'' may be supplied
as New. Given patterns like these in New and Old, the best alignment found by
SP61 is the one shown in Figure \ref{BIRD-TWEETY}.

\begin{figure}
\centering
\begin{BVerbatim}
0 bird      Tweety                                  0
   |          |
1  |   name Tweety #name                            1
   |    |            |
2 bird name        #name canfly wings feathers beak 2

0                          0

1                          1

2 crop lays_eggs ... #bird 2
\end{BVerbatim}
\caption{\small The best alignment found by SP61 with the pattern `bird Tweety'
in New and other patterns in Old as described in the text.} \label{BIRD-TWEETY}
\normalsize
\end{figure}

As before, the inferences which are expressed by this alignment are represented
by the unmatched symbols in the alignment. The fact that Tweety is a bird
allows us to infer that Tweety can fly but it also allows us to infer that
Tweety has wings, feathers and all the other attributes of birds. These
inferences arise directly from the pattern describing the attributes of birds.

There is only one alignment which encodes all the symbols in New. Therefore,
the relative probability of the alignment is 1.0, the relative probability of
`canfly' is 1.0, and likewise for all the other symbols in the alignment, both
those which are matched to New and those which are not.\footnote{At this point
readers may wonder whether the ICMAUS scheme can handle nonmonotonic reasoning:
the fact that additional information about penguins, kiwis and other flightless
birds would invalidate the inference that something being a bird means that it
can fly. Discussion of this point is deferred until Section \ref{non-mon}}

\section{Abductive reasoning}\label{abductive-reason}

In the ICMAUS framework, any subsequence of a pattern may function as what is
`given' in reasoning, with the complementary subsequence functioning as the
inference. Thus, it is just as easy to reason in a `backwards', abductive
manner as it is to reason in a `forwards', deductive manner. We can also reason
from the middle of a pattern outwards, from the ends of a pattern to the
middle, and many other possibilities. In short, the ICMAUS framework allows
seamless integration of probabilistic `deductive' reasoning with abductive
reasoning and other kinds of reasoning which are not commonly recognised.

Figure \ref{TWEETY-CANFLY} shows the best alignment and the other member of its
reference set of alignments which are formed by SP61 with the same patterns in
Old as were used in the example of `deductive' reasoning (Section
\ref{one-step-reasoning}) and with the pattern `Tweety canfly' in New.

By contrast with the example of `deductive' reasoning, there are two alignments
in the reference set of alignments that encode all the symbols in New. These
two alignments represent two alternative sets of abductive inferences that may
be drawn from this combination of New and Old.

\begin{figure}
\centering
\begin{BVerbatim}
0           Tweety       canfly                     0
              |            |
1      name Tweety #name   |                        1
        |            |     |
2 bird name        #name canfly wings feathers beak 2

0                          0

1                          1

2 crop lays_eggs ... #bird 2

(a)

0          Tweety           canfly                       0
             |                |
1     name Tweety #name       |                          1
       |            |         |
2 bat name        #name fur canfly eats_insects ... #bat 2

(b)
\end{BVerbatim}
\caption{\small The best alignment and the other member of its reference set of
alignments which are formed by SP61 with patterns as described in Section
\ref{one-step-reasoning} in Old and `Tweety canfly' in New.}
\label{TWEETY-CANFLY} \normalsize
\end{figure}

With regard to the first alignment (Figure \ref{TWEETY-CANFLY} (a)), `Tweety'
could be a bird with all the attributes of birds, including the ability to fly.
The relative probability of the alignment is 0.8, as is the relative
probability of the pattern for `bird' and every other symbol in that pattern
(apart from the `name' and `\#name' symbols where the relative probability is
1.0).

Alternatively, we may infer from the second alignment (Figure
\ref{TWEETY-CANFLY} (b)) that `Tweety' could be a bat. But in this case the
relative probability of the alignment, the pattern for `bat' and all the
symbols in that pattern (apart from the `name \#name' symbols) is only 0.2.

\section{Reasoning with probabilistic decision networks and decision trees}%
\label{prob-decision-net}

Figure \ref{ENGINE-FAULTS} shows a set of patterns which, in effect, represent
a (highly simplified) decision network for the diagnosis of faults in car
engines. As with other sets of patterns shown in this paper, each pattern has a
notional frequency of occurrence shown in brackets at the end of the pattern.
The patterns supplied to SP61 do not include the English text shown in the
figure.

It should be clear that patterns like the ones shown in Figure
\ref{ENGINE-FAULTS} may be used to represent either a decision network or a
decision tree. The set of patterns in the figure correspond very largely to a
tree structure but, strictly speaking, the structure is a network because
terminal node 12 can be reached via two different paths, and likewise for
terminal node 4.

\begin{figure}
\centering
\begin{BVerbatim}
Start 1 Does the starter turn the engine? (3600)
     1 yes 2 Does the starter turn the engine briskly? (2400)
          2 yes 6 Does the engine start? (1900)
               6 yes 10 Does the engine run smoothly? (700)
                    10 yes 14 Engine is OK. (500)
                    10 no 12 Replace the carburettor. (200)
               6 no 11 Is there fuel in the tank? (1200)
                    11 yes 12 Replace the carburettor. (300)
                    11 no 13 Put fuel in the tank. (900)
          2 no 15 Is the battery flat? (500)
               15 yes 4 Recharge or replace the battery. (400)
               15 no 16 Replace the starter motor. (100)
     1 no 3 Is the battery flat? (1200)
          3 yes 4 Recharge or replace the battery. (800)
          3 no 5 Are there good electrical connections? (400)
               5 yes 8 Are the spark plugs OK? (300)
                    8 yes 17 Replace the starter motor. (100)
                    8 no 18 Replace the spark plugs. (200)
               5 no 9 Repair the electrical connections. (100)
\end{BVerbatim}
\caption{\small A set of patterns representing a highly simplified decision
network for the diagnosis of faults in car engines. As in other examples of
patterns in this paper, each pattern has a notional frequency of occurrence
which is shown in brackets at the end of the pattern. The English text
associated with each pattern is not part of the patterns as they are supplied
to SP61.} \label{ENGINE-FAULTS} \normalsize
\end{figure}

Figure \ref{ENGINE-DIAGNOSIS} shows the best alignment formed by SP61 with the
pattern `Start no no yes no' in New (representing a sequence of yes/no
decisions in the network) and, in Old, the patterns shown in Figure
\ref{ENGINE-FAULTS} (without the English text). The key inference we can make
from this alignment is the numerical identifier `18' at the extreme right of
the figure. This corresponds to the advice ``Replace the spark plugs''.

\begin{figure}
\large
\centering
\begin{BVerbatim}
0 Start   no   no   yes   no    0
    |     |    |     |    |
1 Start 1 |    |     |    |     1
        | |    |     |    |
2       1 no 3 |     |    |     2
             | |     |    |
3            3 no 5  |    |     3
                  |  |    |
4                 5 yes 8 |     4
                        | |
5                       8 no 18 5
\end{BVerbatim}
\normalsize \caption{\small The best alignment formed by SP61 with the pattern
`Start no no yes no' in New and the patterns shown in Figure
\ref{ENGINE-FAULTS} in Old (without the English text).}
\label{ENGINE-DIAGNOSIS} \normalsize
\end{figure}

Since there is only one alignment which matches all the symbols in New, the
probability of the inference in this case is 1.0. If the sequence of symbols in
New is incomplete in some way, e.g., the last symbol (`no') is missing, the
program delivers a set of alternative alignments with probabilities less than
1.0 in much the same way as in the example discussed in Section
\ref{patt-recognition}.

\subsection{So what?}\label{so-what}

Regarding the example in Figure \ref{ENGINE-DIAGNOSIS}, readers may object that
ICMAUS is a long-winded way to achieve something which is done perfectly
adequately with a conventional expert system or even a conventional chart on
paper. Has anything been gained by re-casting the example in an unfamiliar
form?

The main reason for including the example in this article is to show that
multiple alignment as it has been developed in the ICMAUS framework has a much
broader scope than may, at first sight, be assumed. However, one possible
advantage of using the ICMAUS framework is that it is generic for several
different kinds of PR and so can promote the integration of decision networks
and trees with other kinds of PR.

Another possible response to the ``So what?'' question is that the ICMAUS
framework, unlike most conventional discrimination nets or discrimination
trees, does not depend exclusively on input which is both complete and
accurate. It can bridge gaps in information supplied to it (as New) and can
compensate for symbols which have been added or substituted in the input,
provided there are not too many. There is an example with discussion in
\cite{r39}.

\section{Reasoning with `rules'}\label{rules}

The rules in a typical expert system express associations between things in the
form `IF condition THEN consequence (or action)'. As we saw with the example in
Section \ref{simple-example}, we can express an association quite simply as a
pattern like `fire smoke' without the need to make a formal distinction between
the `condition' and the `consequence' or `action'. And, as we saw in Sections 7
and 8, it is possible to use patterns like these quite freely in both a
`forwards' and a `backwards' direction. As was noted in Section
\ref{abductive-reason}, the ICMAUS framework allows inferences to be drawn from
patterns in a totally flexible way: any subsequence of the symbols in a pattern
may function as a condition, with the complementary subsequence as the
corresponding inference.

It is easy to form a chain of inference like ``If A then B, if B then C'' from
patterns like `A B' and `B C'. But if the patterns are `A B' and `C B', the
relative positions of `A' and `C' in the alignment are undefined as described
in Section \ref{one-dimension}. This means that, with the SP52 or SP61 models,
the alignment is treated as being illegal and is discarded. Generalisations of
the models as described in Section \ref{abductive-reason} would probably
provide a solution to this problem but these generalisations have not yet been
attempted.

A way round this problem which can be used with the current models is to adopt
a convention that the symbols in every pattern are arranged in some arbitrary
sequence, e.g., alphabetical, and to include a `framework' pattern and some
additional `service' symbols which together have the effect of ensuring that
every symbol type always has a column to itself. This avoids the kind of
problem described above and allows patterns to be treated as if they were
unordered associations of symbols.

Figure \ref{ARSON-ASSOCIATIONS} shows a small set of patterns representing
well-known associations together with one pattern (`3 destroy 4 10 the\_barn
11') representing the fact that `the barn' has been destroy(ed) and a framework
pattern (`1 2 3 4 5 6 7 8 9 10 11') as mentioned above. In every pattern except
the last, the alphabetic symbols are arranged in alphabetical order. Every
alphabetical symbol has its own slot in the framework, e.g. `black-clouds' has
been assigned to the position between the service symbols `1' and `2'. Every
alphabetical symbol is flanked by the service symbols representing its slot.

\begin{figure}
\centering
\begin{BVerbatim}
4 fire 5 8 smoke 9 (500) 
3 destroy 4 fire 5 (100) 
4 fire 5 matches 6 petrol 7 (300) 
1 black-clouds 2 7 rain 8 (2000) 
1 black-clouds 2 cold 3 9 snow 10 (1000)
3 destroy 4 10 the-barn 11 (1) 
1 2 3 4 5 6 7 8 9 10 11 (7000)
\end{BVerbatim}
\caption{\small Patterns in Old representing well-known associations, together
with one `fact' (`3 destroy 4 10 the-barn 11') and a `framework' pattern (`1 2
3 4 5 6 7 8 9 10 11') as described in the text.} \label{ARSON-ASSOCIATIONS}
\normalsize
\end{figure}

Figure \ref{ARSON-ALIGNMENT} shows the best alignment found by SP61 with the
pattern `accused petrol smoke the-barn' in New and the patterns from Figure
\ref{ARSON-ASSOCIATIONS} in Old. The pattern in New may be taken to represent
the key points in an allegation that an accused person has been seen with
petrol near `the-barn' (which has been destroyed) and that smoke was seen at
the same time.

\begin{figure}
\centering
\fontsize{09.00pt}{10.80pt}
\begin{BVerbatim}
0                       accused    petrol     smoke      the-barn    0
                                     |          |           |
1               4 fire 5 matches 6 petrol 7     |           |        1
                |  |   |         |        |     |           |
2     3 destroy 4 fire 5         |        |     |           |        2
      |    |    |  |   |         |        |     |           |
3 1 2 3    |    4  |   5         6        7 8   |   9 10    |     11 3
      |    |    |  |   |                    |   |   | |     |     |
4     3 destroy |  |   |                    |   |   | 10 the-barn 11 4
                |  |   |                    |   |   |
5               4 fire 5                    8 smoke 9                5
\end{BVerbatim}
\caption{\small The best alignment formed by SP61 with a pattern in New
representing allegations about someone accused of arson and patterns in Old as
shown in Figure \ref{ARSON-ASSOCIATIONS}.} \label{ARSON-ALIGNMENT} \normalsize
\end{figure}

The alleged facts about the accused person do not, in themselves, show that
he/she is guilty of arson. For the jury to find the accused person guilty, they
must understand the connections between the accused person, petrol, smoke and
the destruction of the barn. With this example, the inferences are so simple
(for people) that the prosecuting lawyer would hardly need to spell them out.
But the inferences still need to be made.

The alignment shown in Figure \ref{ARSON-ALIGNMENT} may be interpreted as a
piecing together of the argument that the petrol, with matches or something
similar (that were not seen), was used to start a fire (which was not seen
either), and that the fire explains why smoke was seen and why the barn was
destroyed. Of course, in a more realistic example, there would be many other
clues to the existence of a fire (e.g., charred wood) but the example, as
shown, gives an indication of the way in which evidence and inferences may be
connected together in the ICMAUS paradigm.

\section{Nonmonotonic reasoning and reasoning with default values for variables}\label{non-mon}

The concepts of {\it monotonic} and {\it nonmonotonic} reasoning are well
explained by \cite{r17}. In brief, conventional deductive inference is {\it
monotonic} because, as your set of beliefs grows, so does the set of
conclusions that can be drawn from those beliefs. The deduction that ``Socrates
is mortal'' from the propositions that ``All humans are mortal'' and ``Socrates
is human'' remains true for all time and cannot be invalidated by anything we
learn later. By contrast, the inference that ``Tweety can probably fly'' from
the propositions that ``Most birds fly'' and ``Tweety is a bird'' is {\it
nonmonotonic} because it may be changed if, for example, we learn that Tweety
is a penguin (unless he/she is an astonishing new kind of penguin that can
fly).

This section presents some simple examples which suggest that the ICMAUS
framework may provide a `home' for nonmonotonic reasoning. No attempt is made
to address the several problems associated with nonmonotonic reasoning which
are described in \cite{r17}.

\subsection{Typically, birds fly}\label{birds-fly}

Figure \ref{NON-MON-PATTERNS} shows a set of patterns like the patterns
describing animals that we saw in Figure \ref{TAXONOMY} but adapted to
illustrate nonmonotonic reasoning. The main points to notice in this connection
are that the set of patterns includes one for the class `bird' and one each for
`swallow' and `penguin'. Also, the set of patterns in Figure
\ref{NON-MON-PATTERNS} contains the pattern `bird canfly yes \#canfly \#bird  '.

\begin{figure}
\centering
\begin{BVerbatim}
bird description
          name #name
          structure
               wings #wings
               feathers beak crop
          #structure
          function canfly #canfly lays_eggs #function
     #description #bird (30000)
bird swallow description
                  wings pointed #wings
                  canfly yes #canfly
             #description #swallow #bird (700)
bird penguin description
                  wings stubby #wings
                  canfly no #canfly
             #description #penguin #bird (400)
bird canfly yes #canfly #bird (15000)
name Tweety #name (300)
name John #name(500)
name Tibby #name (400)
\end{BVerbatim}
\caption{\small A set of patterns to illustrate nonmonotonic reasoning.}
\label{NON-MON-PATTERNS} \normalsize
\end{figure}

This last-mentioned pattern provides a `default value' for the variable `canfly
\#canfly' in the class `bird'. The context `bird ... \#bird   ' is included in
the pattern so that the default value applies only to birds and allows for the
possibility that a different default value for the same variable might apply in
the case of, say, insects.

The pattern `bird canfly yes \#canfly \#bird' may be interpreted as a statement
that ``Typically, birds fly''. It should not be interpreted as the
universally-quantified statement that ``All birds fly'' because, as will be
seen below, it can be over-ridden in much the same way as default values in
conventional systems.

Elsewhere \cite{r41a} I have discussed how a universal truth may be expressed
using patterns. In brief, the pattern `bird canfly yes \#canfly \#bird' should be
removed from Old and the pattern defining the class birds should be augmented
so that `canfly \#canfly' within the pattern becomes `canfly yes \#canfly'.

\subsection{Tweety is a bird so, probably, Tweety can fly}\label{tweety-flies}

Figure \ref{NON-MON-ALIGNMENTS-1} shows, at the top, the best alignment found
by SP61 with the pattern `bird Tweety' in New and the patterns from Figure
\ref{NON-MON-PATTERNS} in Old. Below this alignment, in descending order of
CDs, are the other alignments in the reference set. Relative probabilities of
these alignments are shown in the caption to the figure.

The first alignment, which has by far the highest relative probability, tells
us that, as a bird, Tweety almost certainly has wings, feathers, beak and a
crop. In itself, this alignment does not tell us whether or not Tweety can fly.
This accords with a `strict' interpretation of the statement ``Tweety is a
bird'': because of the existence of flightless birds as well as birds that fly,
this statement, in itself, does not tell us whether or not Tweety can fly.

However, the second alignment tells us, without supposing that Tweety is any
particular type of bird, that it is more likely than anything else that Tweety
can fly.

\begin{figure}
\small
\centering
\fontsize{08.00pt}{09.60pt}
\begin{BVerbatim}
0 bd       Twy                                                     0
  |         |
1 bd dn ne  |  #ne se ws #ws fs bk cp #se fn cy #cy ls #fn #dn #bd 1
        |   |   |
2       ne Twy #ne                                                 2

(a)

0 bd       Twy                                                        0
  |         |
1 bd dn ne  |  #ne se ws #ws fs bk cp #se fn cy    #cy ls #fn #dn #bd 1
  |     |   |   |                            |      |              |
2 bd    |   |   |                            cy ys #cy            #bd 2
        |   |   |
3       ne Twy #ne                                                    3

(b)

0 bd          Twy                                            0
  |            |
1 bd sw dn     |         ws pd #ws                 cy ys #cy 1
  |     |      |         |      |                  |      |
2 bd    dn ne  |  #ne se ws    #ws fs bk cp #se fn cy    #cy 2
           |   |   |
3          ne Twy #ne                                        3

0                    0

1        #dn #sw #bd 1
          |       |
2 ls #fn #dn     #bd 2

3                    3

(c)

0 bd          Twy                                            0
  |            |
1 bd pn dn     |         ws sy #ws                 cy no #cy 1
  |     |      |         |      |                  |      |
2 bd    dn ne  |  #ne se ws    #ws fs bk cp #se fn cy    #cy 2
           |   |   |
3          ne Twy #ne                                        3

0                    0

1        #dn #pn #bd 1
          |       |
2 ls #fn #dn     #bd 2

3                    3

(d)
\end{BVerbatim}
\normalsize
\end{figure}

\begin{figure}
\caption{\small {\it (This figure appears on the previous page.)} At the top of
the figure, the best alignment found by SP61 for the pattern `bird Tweety' in
New with the patterns from Figure \ref{NON-MON-PATTERNS} in Old. Below that
alignment, in descending order of CD values, are the other alignments in the
reference set of alignments. {\it Key}: bd = bird, bk = beak, cp = crop, cy =
canfly, dn = description, ee = eagle, fn = function, fs = feathers, ne = name,
ls = lays\_eggs, no = no, pd = pointed, pn = penguin, se = structure, sw =
swallow, sy = stubby, Twy = Tweety, ve = vertebrate, ws = wings, ys = yes. In
order from the top, the relative probabilities of these alignments are: 0.7636,
0.2094, 0.0172, and 0.0098.} \label{NON-MON-ALIGNMENTS-1} \normalsize
\end{figure}

The last alignments in Figure \ref{NON-MON-ALIGNMENTS-1} tell us that, in order
of probability and within the limits imposed by the system's store of
knowledge, Tweety could be a swallow or a penguin. The alignments predict that,
in the first case, Tweety would be able to fly, but not if he/she were a
penguin.

\subsection{Tweety is a penguin, so Tweety cannot fly}\label{tweety-not-fly}

What happens if, when we have learned that Tweety is a bird and inferred that
he/she can probably fly, we are then told that Tweety is a penguin? How, in the
ICMAUS scheme, can we reason nonmonotonically, replacing our first inference
with the new inference that Tweety cannot fly?

We must suppose that the information that Tweety is a bird (`bird Tweety') has
been added to the patterns in Old which are shown in Figure
\ref{NON-MON-PATTERNS}. It is probably not appropriate to add anything to Old
to say that Tweety can probably fly because this is merely inference, not fact.
There may very well be a case for storing tentative inferences, and it seems
likely that people do just that. But the intention in the design of SP61 is
that Old will be restricted to information, in compressed or uncompressed form,
which has, notionally at least, come from New.

Figure \ref{NON-MON-ALIGNMENTS-2} shows the reference set of two alignments
found by SP61 with the pattern `penguin Tweety' in New and with Old augmented
with `bird Tweety' as just described. The first alignment tells us that Tweety
is a bird that cannot fly because he/she is a penguin. The other alignment is
the same but includes the pattern `bird Tweety'. Since both alignments tell us
that Tweety cannot fly, the probability of this conclusion is 1.0, very much
what one would naturally assume from the information that Tweety is a penguin.

Reasoning is nonmonotonic in this example because the previous conclusion that
Tweety could probably fly has been replaced by the conclusion that Tweety
cannot fly.

In this example, the formation of two alignments as shown in Figure
\ref{NON-MON-ALIGNMENTS-1} is somewhat untidy. As we noted in Section
\ref{patt-recognition}, there is probably a case for refining the system so
that it can recognise when one alignment is contained within another. In this
case, alignment (a) is contained within alignment (b) and is, in effect, a
stepping stone in the formation of (b). There is a case for presenting
alignment (b) by itself as the best alignment.

\begin{figure}
\centering
\begin{BVerbatim}
0    pn       Twy                                            0
     |         |
1 bd pn dn     |         ws sy #ws                 cy no #cy 1
  |     |      |         |      |                  |      |
2 bd    dn ne  |  #ne se ws    #ws fs bk cp #se fn cy    #cy 2
           |   |   |
3          ne Twy #ne                                        3

0                    0

1        #dn #pn #bd 1
          |       |
2 ls #fn #dn     #bd 2

3                    3

(a)

0    pn       Twy                                            0
     |         |
1 bd pn dn     |         ws sy #ws                 cy no #cy 1
  |     |      |         |      |                  |      |
2 bd    dn ne  |  #ne se ws    #ws fs bk cp #se fn cy    #cy 2
  |        |   |   |
3 |        ne Twy #ne                                        3
  |            |
4 bd          Twy                                            4

0                    0

1        #dn #pn #bd 1
          |       |
2 ls #fn #dn     #bd 2

3                    3

4                    4

(b)
\end{BVerbatim}
\caption{\small The best alignments found by SP61 with the pattern `penguin
Tweety' in New and with the patterns from Figure \ref{NON-MON-PATTERNS} in Old,
augmented with the pattern `bird Tweety'. The abbreviations of symbols are the
same as in Figure \ref{NON-MON-ALIGNMENTS-1}.} \label{NON-MON-ALIGNMENTS-2}
\normalsize
\end{figure}

\section{Solving geometric analogy problems}\label{geom-analogy}

\begin{figure}
\centering
\includegraphics[width=0.9\textwidth]{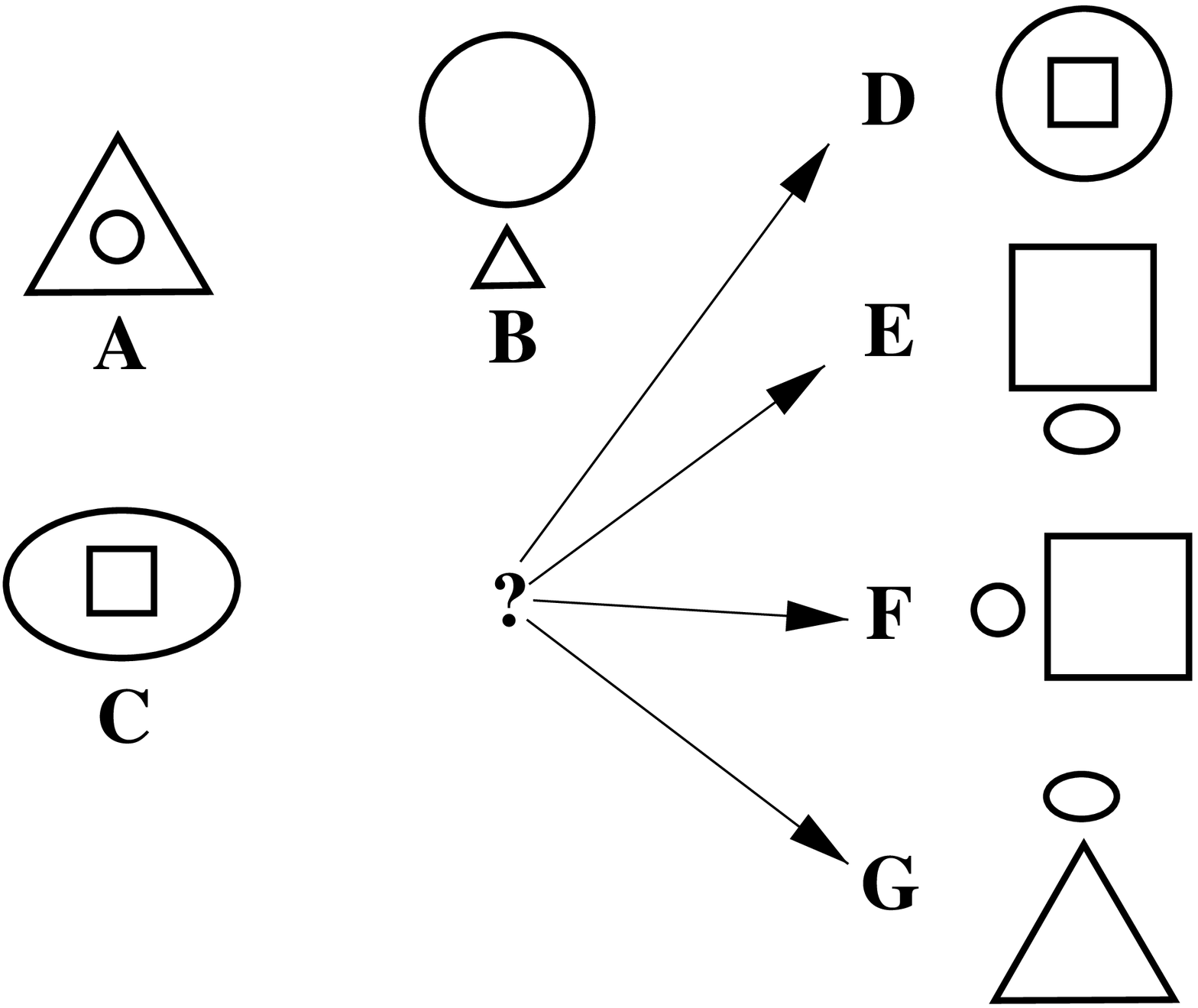}
\caption{\small A geometric analogy problem.}
\label{GEOMETRIC-ANALOGY-PROBLEM}
\end{figure}

Figure \ref{GEOMETRIC-ANALOGY-PROBLEM} shows an example of a well-known type of
simple puzzle - a geometric analogy problem. The task is to complete the
relationship ``A is to B as C is to ?'' using one of the figures `D', `E', `F'
or `G' in the position marked with `?'. For this example, the `correct' answer
is clearly `E'.

What has this got to do with PR? This kind of problem may be seen as an example
of reasoning because it requires a process of ``going beyond the information
given''. Is it probabilistic? In the example shown in Figure
\ref{GEOMETRIC-ANALOGY-PROBLEM}, there seems to be only one `right' answer
which does not seem to leave much room for probabilities less than 1.0. But
with many problems of this type, a case can be made for two or even more
alternative answers and there is a corresponding uncertainty about which answer
is `correct'.

Computer-based methods for solving this kind of problem have existed for some
time (e.g., Evans's well-known heuristic algorithm \cite{r12}). In recent work
\cite{r4, r16}, MLE principles have been applied to good effect. The proposal
here is that, within the general framework of MLE, this kind of problem may be
understood in terms of ICMAUS.

As in some previous work \cite{r4, r16}, the proposed solution assumes that
some mechanism is available which can translate the geometric forms in each
problem into patterns of alpha-numeric symbols like the patterns in other
examples in this article. For example, item `A' in Figure
\ref{GEOMETRIC-ANALOGY-PROBLEM} may be described as `small circle inside large
triangle'.

How this kind of translation may be done is not part of the present proposals
(one such translation mechanism is described in \cite{r12}). As noted elsewhere
\cite{r16}, successful solutions for this kind of problem require consistency
in the way the translation is done. For this example, it would be unhelpful if
item `A' in Figure \ref{GEOMETRIC-ANALOGY-PROBLEM} were described as `large
triangle outside small circle' while item `C' were described as `small square
inside large ellipse'. For any one puzzle, the description needs to stick to
one or other of ``X outside Y'' or ``Y inside X'' - and likewise for
`above/below' and `left-of/right-of'.

Given that the diagrammatic form of the problem has been translated into
patterns as just described, this kind of problem can be cast as a problem of
partial matching, well within the scope of SP61. To do this, symbolic
representations of item A and item B in Figure \ref{GEOMETRIC-ANALOGY-PROBLEM}
are treated as a single pattern, thus:

\begin{center}
\begin{BVerbatim}
A small circle inside large triangle ;
     B large circle above small triangle #,
\end{BVerbatim}
\end{center}

\noindent and this pattern is placed in New. Four other patterns are
constructed by pairing a symbolic representation of item C (on the left) with
symbolic representations of each of D, E, F and G (on the right), thus:

\begin{center}
\begin{BVerbatim}
C1 small square inside large ellipse ;
     D small square inside large circle #
C2 small square inside large ellipse ;
     E large square above small ellipse #
C3 small square inside large ellipse ;
     F small circle left-of large square #
C4 small square inside large ellipse ;
     G small ellipse above large triangle #
\end{BVerbatim}
\end{center}

These four patterns are placed in Old, each with an arbitrary frequency value
of 1.

Figure \ref{GEOMETRIC-ANALOGY-ALIGNMENT} shows the best alignment found by SP61
with New and Old as just described. The alignment is a partial match between
the pattern in New and the second of the four patterns in Old. This corresponds
with the `correct' result (item E) as noted above.

\begin{figure}
\centering
\begin{BVerbatim}
0 A  small circle inside large triangle ; B large circle 0
       |            |      |            |     |
1 C2 small square inside large ellipse  ; E large square 1

0 above small triangle # 0
    |     |            |
1 above small ellipse  # 1
\end{BVerbatim}
\normalsize \caption{\small The best alignment found by SP61 for the patterns
in New and Old as described in the text.} \label{GEOMETRIC-ANALOGY-ALIGNMENT}
\normalsize
\end{figure}

\section{Explaining away `explaining away': ICMAUS as an alternative to Bayesian networks}\label{explaining-away}

In recent years, {\it Bayesian networks} (otherwise known as {\it causal nets},
{\it influence diagrams}, {\it probabilistic networks} and other names) have
become popular as a means of representing probabilistic knowledge and for
probabilistic reasoning (see \cite{r24}).

A Bayesian network is a directed, acyclic graph like the one shown in Figure
\ref{ALARM-BAYESIAN-NETWORK} (below) where each node has zero or more `inputs'
(connections with nodes that can influence the given node) and one or more
`outputs' (connections to other nodes that the given node can influence).

Each node contains a set of conditional probability values, each one the
probability of a given output value for a given input value. With this
information, conditional probabilities of alternative outputs for any node may
be computed for any given {\it combination} of inputs. By combining these
calculations for sequences of nodes, probabilities may be propagated through
the network from one or more `start' nodes to one or more `finishing' nodes.

No attempt will be made in this article to discuss in detail how Bayesian
networks may be modelled in the ICMAUS framework or to compare the two
approaches to probabilistic inference. However, an example is presented below
showing how ICMAUS may provide an alternative to a Bayesian network explanation
of the phenomenon of ``explaining away''.

\subsection{A Bayesian network explanation of ``explaining away''}\label{bayesian-net}

In the words of Judea Pearl \cite[p. 7]{r24}, the phenomenon of `explaining
away' may be characterised as: ``If A implies B, C implies B, and B is true,
then finding that C is true makes A {\it less} credible. In other words,
finding a second explanation for an item of data makes the first explanation
less credible.'' (his italics). Here is an example described by \cite[pp.
8-9]{r24}:

\begin{quote}
{\it Normally an alarm sound alerts us to the possibility of a burglary. If
somebody calls you at the office and tells you that your alarm went off, you
will surely rush home in a hurry, even though there could be other causes for
the alarm sound. If you hear a radio announcement that there was an earthquake
nearby, and if the last false alarm you recall was triggered by an earthquake,
then your certainty of a burglary will diminish.}
\end{quote}

Although it is not normally presented as an example of nonmonotonic reasoning,
this kind of effect in the way we react to new information is similar to the
example we considered in Section \ref{non-mon} because new information has an
impact on inferences that we formed on the basis of information that was
available earlier.

\begin{figure}
\centering
\includegraphics[width=0.9\textwidth]{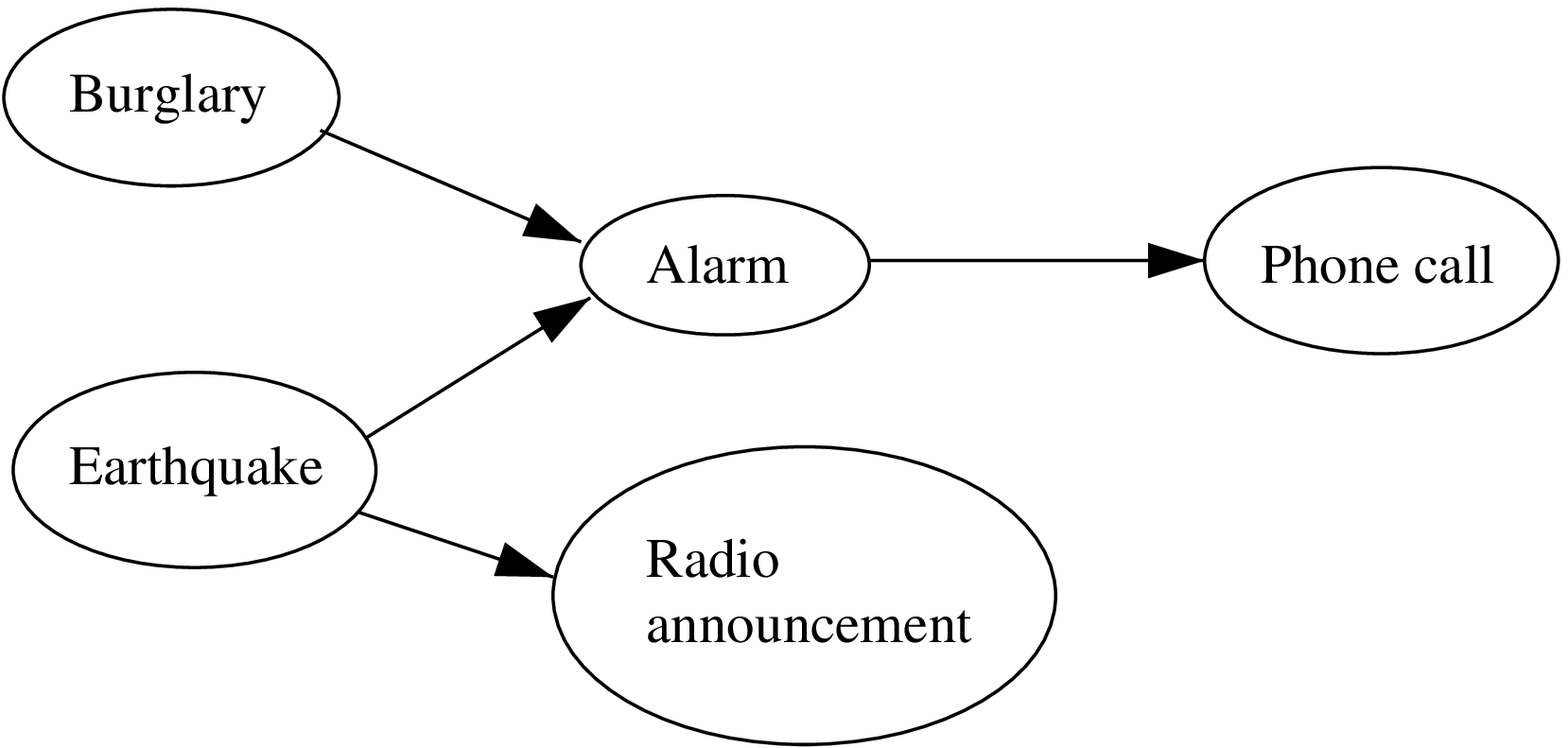}
\caption{A Bayesian network representing causal relationships discussed in the
text. In this diagram, ``Phone call'' means ``a phone call about the alarm
going off'' and ``Radio announcement'' means ``a radio announcement about an
earthquake''.} \label{ALARM-BAYESIAN-NETWORK}
\end{figure}

The causal relationships in the example just described may be captured in a
Bayesian network like the one shown in Figure \ref{ALARM-BAYESIAN-NETWORK}.

\cite{r24} argues that, with appropriate values for conditional probability,
the phenomenon of ``explaining away'' can be explained in terms of this network
(representing the case where there is a radio announcement of an earthquake)
compared with the same network without the node for "radio announcement"
(representing the situation where there is no radio announcement of an
earthquake).

\subsection{Representing contingencies with patterns and frequencies}\label{contingencies}

To see how this phenomenon may be understood in terms of ICMAUS, consider,
first, the set of patterns shown in Figure \ref{ALARM-PATTERNS}, which are to
be stored in Old. The first four patterns in the figure show events which occur
together in some notional sample of the `World' together with their frequencies
of occurrence in the sample.

Like other knowledge-based systems, an ICMAUS system would normally be used
with a `closed-world' assumption that, for some particular domain, the
knowledge stored in the knowledge base is comprehensive. Thus, for example, a
travel booking clerk using a database of all flights between cities will assume
that, if no flight is shown between, say, Edinburgh and Paris, then no such
flight exists. Of course, the domain may be only `flights provided by one
particular airline', in which case the booking clerk would need to check
databases for other airlines. In systems like Prolog, the closed-world
assumption is the basis of `negation as failure': if a proposition cannot be
proved with the clauses provided in a Prolog program then, in terms of that
store of knowledge, the proposition is assumed to be false.

In the present case, we shall assume that the closed-world assumption applies
so that the absence of any pattern may be taken to mean that the corresponding
pattern of events did not occur, at least not with a frequency greater than one
would expect by chance.

\begin{figure}
\centering
\begin{BVerbatim}
alarm phone_alarm_call (980)
earthquake alarm (20) 
earthquake radio_earthquake_announcement (40)
burglary alarm (1000)
e1 earthquake e2 (40)
\end{BVerbatim}
\caption{\small A set of patterns to be stored in Old in an example of
`explaining away'. The symbol `phone\_alarm\_call' is intended to represent a
phone call conveying news that the alarm sounded;
`radio\_earthquake\_announcement' represents an announcement on the radio that
there has been an earthquake. The symbols `e1' and `e2' represent other
contexts for `earthquake' besides the contexts `alarm' and
`radio\_earthquake\_announcement'.} \label{ALARM-PATTERNS} \normalsize
\end{figure}

The fourth pattern shows that there were 1000 occasions when there was a
burglary and the alarm went off and the second pattern shows just 20 occasions
when there was an earthquake and the alarm went off (presumably triggered by
the earthquake). Thus we have assumed that burglaries are much more common than
earthquakes. Since there is no pattern showing the simultaneous occurrence of
an earthquake, burglary and alarm, we shall infer from the closed-world
assumption that this constellation of events was not recorded during the
sampling period.

The first pattern shows that, out of the 1020 cases when the alarm went off,
there were 980 cases where a telephone call about the alarm was made. Since
there is no pattern showing telephone calls (about the alarm) in any other
context, the closed-world assumption allows us to assume that there were no
false positives (including hoaxes): telephone calls about the alarm when no
alarm had sounded.

Some of the frequencies shown in Figure \ref{ALARM-PATTERNS} are intended to
reflect the two probabilities suggested for this example in \cite[p. 49]{r24}:
"... the [alarm] is sensitive to earthquakes and can be accidentally (P = 0.20)
triggered by one. ... if an earthquake had occurred, it surely (P = 0.40) would
be on the [radio] news.''

In our example, the frequency of `earthquake alarm' is 20, the frequency of
`earthquake radio\_earthquake\_announcement' is 40 and the frequency of
`earthquake' in other contexts is 40. Since there is no pattern like
`earthquake alarm radio\_earthquake\_announcement' or `earthquake
radio\_earthquake\_announcement alarm' representing cases where an earthquake
triggers the alarm and also leads to a radio announcement, we may assume that
cases of that kind have not occurred. As before, this assumption is based on
the closed-world assumption that the set of patterns is a reasonably
comprehensive representation of non-random associations in this small world.

The pattern at the bottom, with its frequency, shows that an earthquake has
occurred on 40 occasions in contexts where the alarm did not ring and there was
no radio announcement.

\subsection{Approximating the temporal order of events}\label{temporal-order}

In these patterns and in the alignments shown below, the left-to-right order of
symbols may be regarded as an approximation to the order of events in time.
Thus, in the first pattern, `phone\_alarm\_call' (a phone call to say the alarm
has gone off) follows `alarm' (the alarm itself); in the second pattern,
`alarm' follows `earthquake' (the earthquake which, we may guess, triggered the
alarm); and so on. A single dimension can only approximate the order of events
in time because it cannot represent events which overlap in time or which occur
simultaneously. However, this kind of approximation has little or no bearing on
the points to be illustrated here.

\subsection{Other considerations}\label{other-things}

Other points relating to the patterns shown in Figure \ref{ALARM-PATTERNS}
include:

\begin{itemize}
\item No attempt has been made to represent the idea that ``the last false
alarm you recall was triggered by an earthquake'' \cite[p. 9]{r24}. At some
stage in the development of the SP system, an attempt may be made to take
account of recency. \item With these imaginary frequency values, it has been
assumed that burglaries (with a total frequency of occurrence of 1160) are much
more common than earthquakes (with a total frequency of 100). As we shall see,
this difference reinforces the belief that there has been a burglary when it is
known that the alarm has gone off (but without additional knowledge of an
earthquake). \item In accordance with Pearl's example \cite[p. 49]{r24} (but
contrary to the phenomenon of looting during earthquakes), it has been assumed
that earthquakes and burglaries are independent. If there was some association
between them, then, in accordance with the closed-world assumption, there
should be a pattern in Figure \ref{ALARM-PATTERNS} representing the
association.
\end{itemize}

\subsection{Formation of alignments: the burglar alarm has sounded}\label{burglar-alarm}

Receiving a phone call to say that one's house alarm has gone off may be
represented by placing the symbol `phone\_alarm\_call' in New. Figure
\ref{ALARM-ALIGNMENTS-1} shows, at the top, the best alignment formed by SP61
in this case with the patterns from Figure \ref{ALARM-PATTERNS} in Old. The
other two alignments in the reference set are shown below the best alignment,
in order of CD value and relative probability. The actual values for CD and
relative probability are given in the caption to Figure \ref{ALARM-PATTERNS}.

\begin{figure}
\centering
\begin{BVerbatim}
0       phone_alarm_call 0
               |
1 alarm phone_alarm_call 1

(a)

0                phone_alarm_call 0
                        |
1          alarm phone_alarm_call 1
             |
2 burglary alarm                  2

(b)

0                  phone_alarm_call 0
                          |
1            alarm phone_alarm_call 1
               |
2 earthquake alarm                  2

(c)
\end{BVerbatim}
\caption{The best alignment (at the top) and the other three alignments in its
reference set formed by SP61 with the pattern `phone\_alarm\_call' in New and
the patterns from Figure \ref{ALARM-PATTERNS} in Old. In order from the top,
the values for CD with relative probabilities in brackets are: 19.91 (0.6563),
18.91 (0.3281), 14.52 (0.0156).} \label{ALARM-ALIGNMENTS-1}
\end{figure}

The unmatched symbols in these alignments represent inferences made by the
system. The probabilities for these symbols which are calculated by SP61 (using
the method described in Section \ref{prob-reason}) are shown in Table
\ref{SYMBOL-PROBABILITIES}. These probabilities do not add up to 1 and we
should not expect them to because any given alignment can contain two or more
of these symbols.

The most probable inference is the rather trivial inference that the alarm has
indeed sounded. This reflects the fact that there is no pattern in Figure
\ref{ALARM-PATTERNS} representing false positives for telephone calls about the
alarm. Apart from the inference that the alarm has sounded, the most probable
inference (p = 0.3281) is that there has been a burglary. However, there is a
distinct possibility that there has been an earthquake - but the probability in
this case (p = 0.0156) is much lower than the probability of a burglary.

\begin{table}
\centering
\begin{tabular}{ll}
\it Symbol & \it Probability \\
\\
alarm & 1.0 \\
burglary & 0.3281 \\
earthquake & 0.0156 \\
\end{tabular}
\caption{\small The probabilities of unmatched symbols, calculated by SP61 for
the four alignments shown in Figure \ref{ALARM-ALIGNMENTS-1}. The probability
of `phone\_alarm\_call' is 1.0 because it is supplied as a `fact' in New.}
\label{SYMBOL-PROBABILITIES} \normalsize
\end{table}

These inferences and their relative probabilities seem to accord quite well
with what one would naturally think following a telephone call to say that the
burglar alarm at one's house has gone off (given that one was living in a part
of the world where earthquakes were not vanishingly rare).

\subsection{Formation of alignments: the burglar alarm has sounded and there is a radio announcement of an earthquake}\label{radio-announcement}

In this example, the phenomenon of `explaining away' occurs when you learn not
only that the burglar alarm has sounded but that there has been an announcement
on the radio that there has been an earthquake. In terms of the ICMAUS model,
the two events (the phone call about the alarm and the announcement about the
earthquake) can be represented in New by a pattern like this:

\begin{center}
\begin{BVerbatim}
`phone_alarm_call radio_earthquake_announcement'
\end{BVerbatim}
\end{center}

or `radio\_earthquake\_announcement phone\_alarm\_call'. The order of the two
symbols does not matter because it makes no difference to the result, except
for the order in which columns appear in the best alignment.

\begin{figure}
\centering
\fontsize{08.00pt}{09.60pt}
\begin{BVerbatim}
0                  phone_alarm_call radio_earthquake_announcement 0
                          |                       |
1            alarm phone_alarm_call               |               1
               |                                  |
2 earthquake alarm                                |               2
      |                                           |
3 earthquake                        radio_earthquake_announcement 3

(a)

0 phone_alarm_call radio_earthquake_announcement 0
                                 |
1 earthquake       radio_earthquake_announcement 1

(b)

0       phone_alarm_call radio_earthquake_announcement 0
               |
1 alarm phone_alarm_call                               1

(c)

0                phone_alarm_call radio_earthquake_announcement 0
                        |
1          alarm phone_alarm_call                               1
             |
2 burglary alarm                                                2

(d)

0                  phone_alarm_call radio_earthquake_announcement 0
                          |
1            alarm phone_alarm_call                               1
               |
2 earthquake alarm                                                2

(e)
\end{BVerbatim}
\caption{\small At the top, the best alignment formed by SP61 with the pattern
`phone\_alarm\_call radio\_earthquake\_announcement' in New and the patterns
from Figure \ref{ALARM-PATTERNS} in Old. Other alignments formed by SP61 are
shown below. From the top, the CD values are: 74.64, 54.72, 19.92, 18.92, and
14.52.} \label{ALARM-ALIGNMENTS-2} \normalsize
\end{figure}

In this case, there is only one alignment (shown at the top of Figure
\ref{ALARM-ALIGNMENTS-2}) that can `explain' all the information in New. Since
there is only this one alignment in the reference set for the best alignment,
the associated probabilities of the inferences that can be read from the
alignment (`alarm   ' and `earthquake') are 1.0.

These results show broadly how `explaining away' may be explained in terms of
ICMAUS. The main point is that the alignment or alignments that provide the
best `explanation' of a telephone call to say that one's burglar alarm has
sounded is different from the alignment or alignments that best explain the
same telephone call coupled with an announcement on the radio that there has
been an earthquake. In the latter case, the best explanation is that the
earthquake triggered the alarm. Other possible explanations have lower
probabilities.

\subsection{Other possible alignments}\label{other-alignments}

The foregoing account of `explaining away' in terms of ICMAUS is not entirely
satisfactory because it does not say enough about alternative explanations of
what has been observed. This subsection tries to plug this gap.

What is missing from the account of `explaining away' in the previous
subsection is any consideration of such other possibilities as, for example:

\begin{itemize}
\item A burglary (which triggered the alarm) and, at the same time, an
earthquake (which led to a radio announcement), or \item An earthquake that
triggered the alarm and led to a radio announcement and, at the same time, a
burglary that did not trigger the alarm. \item And many other unlikely
possibilities of a similar kind.
\end{itemize}

Alternatives of this kind may be created by combining alignments shown in
Figure \ref{ALARM-ALIGNMENTS-2} with each other, or with patterns or symbols
from Old, or both these things. The two examples just mentioned are shown in
Figure \ref{ALARM-ALIGNMENTS-3}.

\begin{figure}
\centering
\fontsize{09.00pt}{10.80pt}
\begin{BVerbatim}
0                phone_alarm_call radio_earthquake_announcement 0
                        |                       |
1          alarm phone_alarm_call               |               1
             |                                  |
2 burglary alarm                                |               2
                                                |
3 earthquake                      radio_earthquake_announcement 3

(a)

0                  phone_alarm_call radio_earthquake_announcement 0
                          |                       |
1            alarm phone_alarm_call               |               1
               |                                  |
2 earthquake alarm                                |               2
      |                                           |
3 earthquake                        radio_earthquake_announcement 3

4 burglary                                                        4

(b)
\end{BVerbatim}
\caption{Two alignments discussed in the text. (a) An alignment created by
combining the second and fourth alignment from Figure \ref{ALARM-ALIGNMENTS-2}.
CD = 73.64, Absolute P = 5.5391e-5. (b) An alignment created from the first
alignment in Figure \ref{ALARM-ALIGNMENTS-2} and the symbol `burglary'. CD =
72.57, Absolute P = 2.6384e-5.} \label{ALARM-ALIGNMENTS-3}
\end{figure}

Any alignment created by combining alignments as just described may be
evaluated in exactly the same way as the alignments formed directly by SP61.
CDs and absolute probabilities for the two example alignments are shown in the
caption to Figure \ref{ALARM-ALIGNMENTS-3}.

Given the existence of alignments like those shown in Figure
\ref{ALARM-ALIGNMENTS-3}, values for relative probabilities of alignments will
change. The best alignment from Figure \ref{ALARM-ALIGNMENTS-2} and the two
alignments from Figure \ref{ALARM-ALIGNMENTS-3} constitute a reference set
because they all `encode' the same symbols from New. However, there are
probably several other alignments that one could construct that would belong in
the same reference set.

Given a reference set containing the first alignment in Figure
\ref{ALARM-ALIGNMENTS-2} and the two alignments in Figure
\ref{ALARM-ALIGNMENTS-3}, values for relative probabilities are shown in Table
\ref{ABS-REL-PROBS}, together with the absolute probabilities from which they
were derived. Whichever measure is used, the alignment which was originally
judged to represent the best interpretation of the available facts has not been
dislodged from this position.

\begin{table}
\centering
\begin{tabular}{lll}
\it Alignment & \it Absolute & \it Relative \\
 & \it probability & \it probability \\
\\
(a) in Figure \ref{ALARM-ALIGNMENTS-2} & 1.1052e-4 & 0.5775 \\
(a) in Figure \ref{ALARM-ALIGNMENTS-3} & 5.5391e-5 & 0.2881 \\
(b) in Figure \ref{ALARM-ALIGNMENTS-3} & 2.6384e-5 & 0.1372 \\
\end{tabular}
\caption{\small Values for absolute and relative probability for the best
alignment in Figure \ref{ALARM-ALIGNMENTS-2} and the two alignments in Figure
\ref{ALARM-ALIGNMENTS-3}.} \label{ABS-REL-PROBS}
\end{table}

\section{Conclusion}\label{conclusion}

In this article, I have outlined the concept of {\it information compression by
multiple alignment, unification and search} as it has been developed in this
research program and, with examples, have tried to show how the ICMAUS paradigm
may provide a framework within which various kinds of probabilistic reasoning
may be understood. Substantially more detail about these ideas may be found in
\cite{r38, r39, r40}.

This approach to understanding probabilistic reasoning seems to be sufficiently
promising to merit further investigation and development.

\section*{Acknowledgements}

\small I am grateful to Prof. C. S Wallace of Monash University, to Prof. Ray
Solomonoff of Oxbridge Research Inc., to Prof. Tim Porter of the Department of
Mathematics, University of Wales, Bangor (UWB) and to Mr. John Hornsby of the
School of Electronic Engineering and Computer Systems, UWB, for useful
comments, suggestions and discussions of various ideas presented in this
article. 

\end{document}